\documentclass[11pt]{article}

\usepackage[utf8]{inputenc}
\usepackage[T1]{fontenc}
\usepackage{amsmath,amssymb,amsthm}
\usepackage{graphicx}
\usepackage{booktabs}
\usepackage{hyperref}
\usepackage{cleveref}
\usepackage{natbib}
\usepackage{microtype}
\usepackage[margin=1in]{geometry}
\usepackage{xcolor}
\usepackage{enumitem}
\usepackage{placeins}   
\graphicspath{{figures/}}

\newcommand{\eg}{\textit{e.g.}}

\title{Tracking vs.\ Deciding: The Dual-Capability Bottleneck\\in Searchless Chess Transformers}

\author{
  \textbf{Quanhao Li} \\
  Abbey Park High School \\
  \texttt{heinzliyuanhao@gmail.com}
  \and
  \textbf{Wei Jiang} \\
  Shanghai Soong Ching Ling School \\
  \texttt{chesssolveasy@gmail.com}
}

\date{March 2026}

\begin{document}
\maketitle

\begin{abstract}
A human-like chess engine should mimic the style, errors, and consistency of a strong human player rather than maximize playing strength.
We show that training such an engine from move sequences alone forces the model to learn two capabilities simultaneously---\emph{state tracking} (reconstructing the board from the move history) and \emph{decision quality} (choosing good moves given that knowledge)---and that these capabilities impose contradictory demands on training data.
Low-rated games supply the positional diversity essential for tracking; high-rated games supply the quality signal for decision learning.
Filtering out low-rated data, the intuitive route to better moves, consistently degrades performance: illegal-move rates double and middlegame accuracy collapses.

We formalize this tension as a \emph{dual-capability bottleneck}, $P \le \min(T, Q)$, where performance is limited by whichever capability is weaker.
Guided by this framework, we (i)~scale the model from 28\,M to 120\,M parameters to repair tracking, then (ii)~introduce \emph{Elo-weighted training}---reducing the gradient contribution of low-rated games without removing them---to boost decision quality while preserving diversity.
A complete $2 \times 2$ factorial ablation supports the view that scaling primarily fixes tracking while weighting primarily improves decisions, and their combination produces the strongest practical-play gains.
A weighting sweet spot exists: linear weighting ($r{=}20$) achieves the optimal balance, while exponential weighting ($r{=}200$) destroys tracking despite achieving lower validation loss.
We further introduce a \emph{coverage-decay formula}, $t^{*} = \log(N/k_{\mathrm{crit}})/\log b$, as a characteristic reliability horizon for intra-game degeneration risk.

Our final model (120\,M parameters, no search) reaches Lichess bullet \textbf{2570} over 253 rated games---to our knowledge, the strongest reported pure move-sequence chess model (top~1\% of Lichess bullet players).
Under native-input evaluation, it achieves 55.2\% Top-1 accuracy on human move prediction, compared to 50.0\% (Maia-2 rapid) and 50.4\% (Maia-2 blitz); part of this advantage reflects the richer sequence input rather than model quality alone.
The model plays on Lichess without triggering the platform's anti-cheating system, consistent with human-like move and timing patterns.
Unlike position-based methods, the sequence input naturally encodes full game history, enabling history-dependent decision-making that single-position models structurally cannot exhibit.
\end{abstract}

\section{Introduction}\label{sec:intro}

\subsection{The goal}\label{sec:goal}

Superhuman chess engines have existed for over two decades, yet for most players they are practically useless as training partners.
When a 1500-rated player asks Stockfish why their move was wrong, the engine suggests a line involving a quiet rook maneuver whose purpose only becomes clear twelve moves later---a sequence no human below master level could conceive, let alone execute.
The gap between superhuman play and human understanding is not a feature; it is a barrier.

What players need is a \emph{relatable} opponent: an engine that plays like a human slightly above their level---stylistically coherent across an entire game, capable of occasional inaccuracies that mirror real human tendencies, yet never blundering material through sheer positional confusion.
The goal is not to maximize playing strength but to maximize \emph{human-likeness} at a target skill level.

Prior work has made significant progress.
Maia and Maia-2 \citep{mcilroy2020aligning,tang2024maia2} predict human moves with impressive accuracy by training on board positions at various skill levels.
However, these models operate on single board states (FEN) and have no access to game history---they cannot distinguish the same position reached via an aggressive gambit from one reached via quiet maneuvering, and therefore cannot maintain stylistic coherence across a game.
At the other end, attempts to repurpose general-purpose LLMs for chess \citep{feng2023chessgpt,zhang2025chessgames} treat move sequences as text but achieve only modest playing strength (${\sim}1788$ Elo), far below strong amateur level.

In this work, we build a searchless autoregressive Transformer that learns exclusively from move sequences---with no board representation, no search, and no hand-crafted chess knowledge---and plays bullet chess at the level of a 2500+ rated human.

\subsection{Why this is hard: the tracking--decision contradiction}\label{sec:contradiction}

Our model is a standard decoder-only Transformer whose input is a sequence of moves in UCI notation; it predicts the next move via next-token prediction.
Crucially, it never sees the board.
Every piece's location must be implicitly reconstructed from the move history---a 64-square board state inferred from a growing sequence of algebraic tokens.

This places two simultaneous demands on the model:

\paragraph{State tracking ($T$).}
The model must maintain an accurate internal representation of the current board position---which pieces occupy which squares, what has been captured, whether castling rights remain.
Tracking failure manifests as illegal moves: attempts to move pieces that are not where the model believes them to be.

\paragraph{Decision quality ($Q$).}
Given an accurate board representation, the model must select moves consistent with human-level positional and tactical understanding.
We measure decision quality through centipawn loss (CPL) and move-prediction accuracy on high-rated subsets.

The central difficulty is that these two capabilities require \emph{contradictory} training data:
\begin{itemize}[nosep]
  \item \textbf{Tracking} is trained by \emph{diversity} of positions. Low-rated players make unconventional moves that produce unusual piece configurations rarely seen in high-level play. These ``messy'' positions are the training signal the model needs for robust board reconstruction.
  \item \textbf{Decision quality} is trained by \emph{quality} of moves. High-rated players choose principled moves that embody sound positional understanding.
\end{itemize}

The naive approach---filtering out low-rated games---consistently fails: illegal-move rates double, middlegame and endgame accuracy collapse, and the filtered model loses decisively to the unfiltered baseline in head-to-head play (\Cref{sec:paradox}).
A single tracking failure (blundering a piece) is catastrophically more costly than any number of slightly suboptimal but legal moves.

\begin{figure}[htbp]
\centering
\includegraphics[width=\columnwidth]{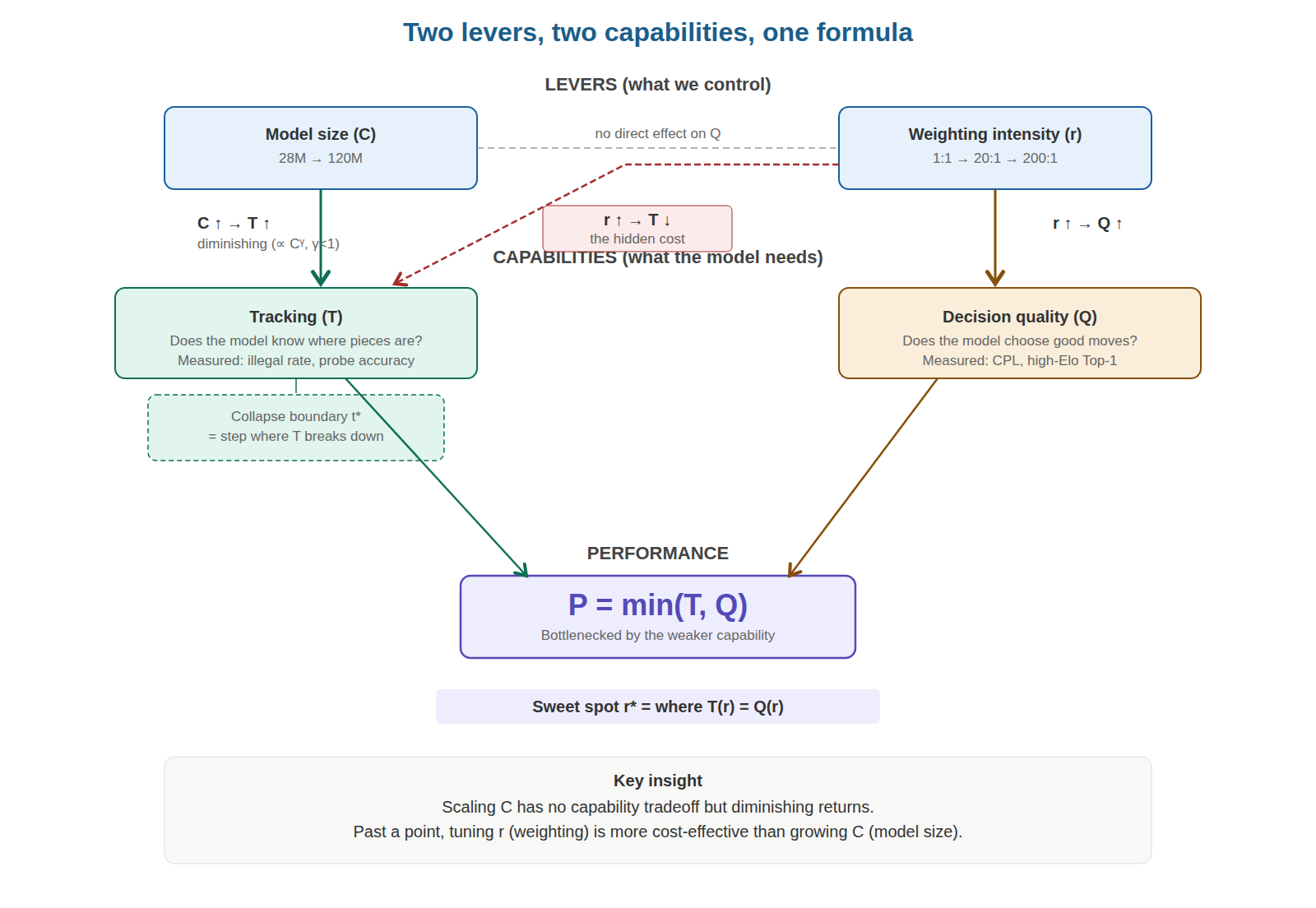}
\caption{Core variable relationships. Two controllable levers (model capacity and data weighting) feed two capabilities (tracking $T$ and decision quality $Q$), governed by the bottleneck $P \le \min(T, Q)$.}
\label{fig:variables}
\end{figure}

We formalize this as a \emph{bottleneck}:
\begin{equation}\label{eq:bottleneck}
  P \;\le\; \min(T,\, Q),
\end{equation}
where $P$ is effective playing strength.
Performance is limited by whichever capability is weaker.
\Cref{fig:variables} summarizes the complete variable structure.

\subsection{A chain of failures leading to a solution}\label{sec:chain}

Our experimental path was driven by successive failure diagnoses (\Cref{fig:chain}):

\begin{figure}[htbp]
\centering
\includegraphics[width=\columnwidth]{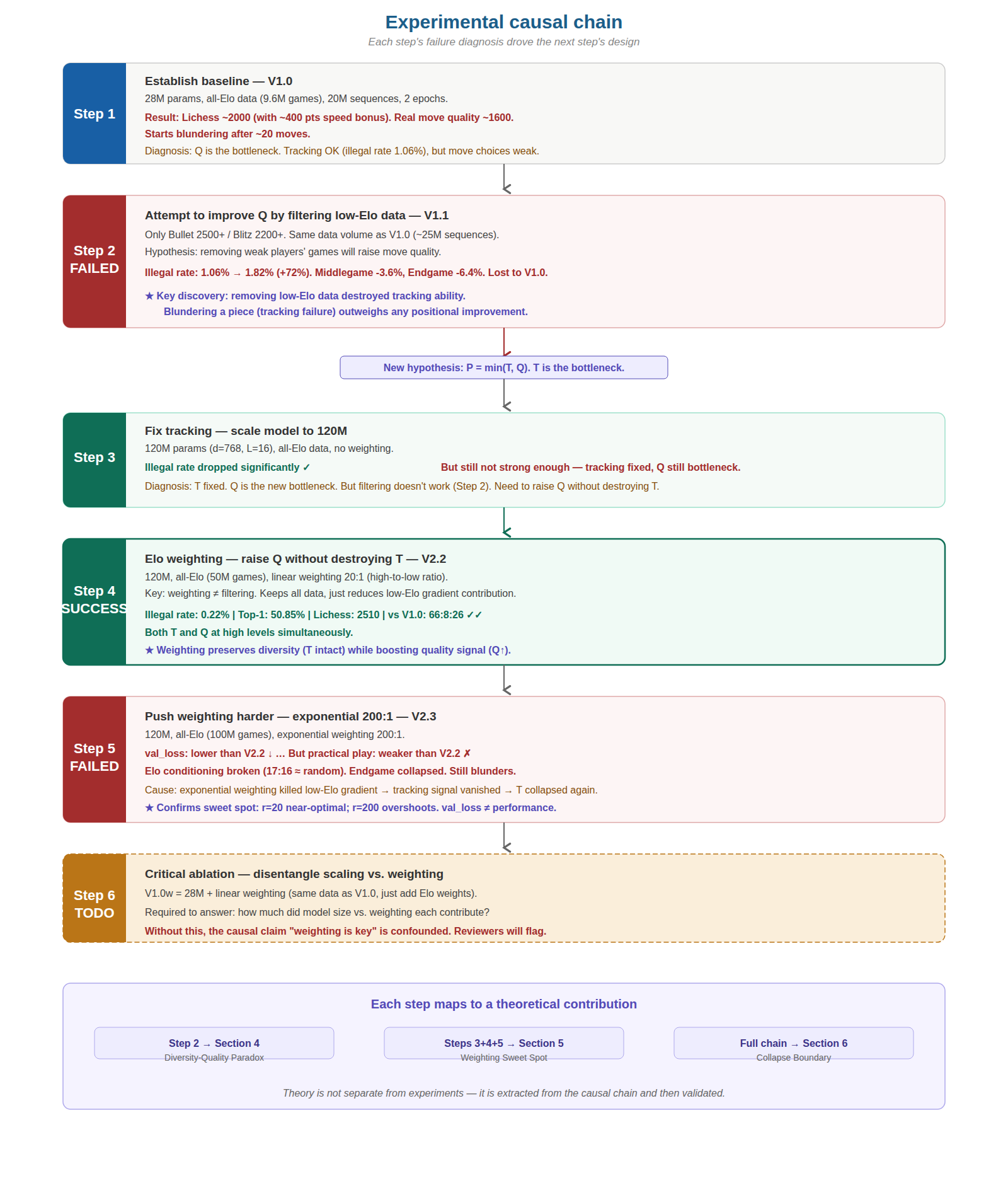}
\caption{Experimental causal chain. Each step's failure diagnosis motivates the next intervention, from baseline through filtering failure, model scaling, Elo weighting, and the discovery of the weighting sweet spot.}
\label{fig:chain}
\end{figure}

\begin{enumerate}[nosep]
  \item \textbf{Baseline} (28\,M, all Elo): ${\sim}2000$ Lichess bullet, but quality degrades as games progress. Illegal rate: 1.06\%.
  \item \textbf{Filtering} (28\,M, high Elo only): illegal rate nearly doubles to 1.82\%; middlegame $-3.6$\,pp, endgame $-6.4$\,pp. Loses to baseline. \emph{Diagnosis: $T$ crashed, became new bottleneck.}
  \item \textbf{Scaling} (28\,M $\to$ 120\,M, uniform): illegal rate drops to 0.22\%, Top-1 rises to 50.85\%, but only modest head-to-head gains (80--54 over 200 games, $p{=}0.03$). \emph{Diagnosis: $T$ fixed, $Q$ still bottlenecked.}
  \item \textbf{Elo-weighted training} (120\,M, linear $r{=}20$): slightly higher Top-1 (51.2\% vs.\ 50.85\%) at a small tracking cost (0.26\% vs.\ 0.22\% illegal), but dominates V2.0 41--24 and baseline 66--8--26. Lichess bullet \textbf{2570}. \emph{Sweet spot reached.}
  \item \textbf{Over-weighting} (120\,M, exponential $r{=}200$): lower validation loss but weaker play; endgame collapses, Elo conditioning breaks. \emph{Diagnosis: $T$ re-emerges as bottleneck.}
\end{enumerate}

\subsection{Contributions}\label{sec:contributions}

\begin{enumerate}[nosep]
  \item \textbf{The dual-capability bottleneck.} We identify and formalize the tracking--decision tradeoff: $P \le \min(T, Q)$, explaining why filtering low-quality data consistently degrades sequence-based chess models (\Cref{sec:paradox}).

  \item \textbf{Elo-weighted training and full factorial ablation.} We propose Elo-weighted loss as a principled resolution: it boosts $Q$ while preserving the diversity that $T$ requires. A complete $2 \times 2$ factorial (two scales $\times$ two weighting regimes) supports the view that scaling primarily fixes tracking while weighting primarily improves decisions, and their combination produces the strongest practical-play gains (\Cref{sec:sweetspot}).

  \item \textbf{Intra-game degeneration analysis.} We introduce per-game degeneration points and the coverage-decay horizon $t^{*} = \log(N/k_{\mathrm{crit}})/\log b$, showing that degeneration is localized within games and delayed by stronger models (\Cref{sec:degeneration}).

  \item \textbf{A competitive human-like engine from pure move sequences.} Our final model (120\,M, no search) reaches Lichess bullet 2570 (top~1\%), outperforms Maia-2 under native-input evaluation by 5\,pp on human move prediction, shows stronger alignment with human error patterns, and exhibits history-dependent decision-making that FEN-only models structurally cannot exhibit (\Cref{sec:humanlike}).
\end{enumerate}

\FloatBarrier
\section{Related Work}\label{sec:related}

\subsection{Superhuman chess engines}

The dominant paradigm combines neural evaluation with explicit tree search.
AlphaZero \citep{silver2018general} and Leela Chess Zero \citep{lczero2026} use convolutional networks with MCTS, trained via self-play reinforcement learning.
Stockfish \citep{stockfish2023} pairs NNUE evaluation with alpha-beta search.
These systems achieve superhuman strength but are fundamentally unsuitable for human-like behavior: they model how to win, not how humans think.

A notable departure is \citet{ruoss2024amortized}, who train a 270\,M Transformer to predict Stockfish action-values from FEN, achieving Lichess blitz 2895 without search.
However, this is a distillation of Stockfish---its errors bear no resemblance to human errors, and its FEN-only input encodes no game history.

\subsection{Human-like chess engines}

Maia \citep{mcilroy2020aligning} pioneered human-move prediction by training separate Leela-architecture models for each Elo bracket.
Maia-2 \citep{tang2024maia2} introduced a skill-aware attention mechanism on a ResNet backbone, producing a single unified model conditioned on player Elo.

Two limitations of the Maia line are relevant here.
First, both operate on single board positions (FEN): two games reaching the same position via different paths produce identical predictions, precluding game-level style coherence.
Second, neither investigates \emph{data composition}---whether to include or exclude low-rated games, or how the quality--diversity balance affects model capabilities.
As we show, these questions are critical.

\subsection{Sequence models and implicit state tracking}

Othello-GPT \citep{li2023emergent} demonstrated that a GPT trained on move sequences develops an internal world model, confirmed by linear probes on hidden states.
\citet{toshniwal2022chess} systematically studied the conditions under which Transformers learn state tracking from chess move sequences, finding that full attention and sufficient data volume are necessary.
However, they did not investigate how \emph{data composition}---the balance between low-rated and high-rated games---affects tracking quality, nor did they identify any tension between tracking and decision capabilities.
\citet{karvonen2024emergent} extended the emergent world-model finding to chess, showing that a 50\,M GPT trained on PGN sequences learns 64-square board representations and implicit Elo estimation, achieving ${\sim}1500$ Elo.

ChessGPT \citep{feng2023chessgpt} and ChessLLM \citep{zhang2025chessgames} fine-tune general-purpose LLMs on chess data, with ChessLLM reaching ${\sim}1788$ Elo.
General-purpose LLMs dedicate most parameters to linguistic capabilities, leaving limited effective capacity for board tracking.
Neither paper analyzes why performance plateaus or investigates training data composition.

To our knowledge, our work is the first to (i)~demonstrate that a dedicated sequence model can reach strong human-level play (2500+), and (ii)~systematically study how data composition affects tracking and decision quality.

\FloatBarrier
\section{Method}\label{sec:method}

\subsection{Architecture and input representation}\label{sec:arch}

We use a standard decoder-only Transformer with Pre-RMSNorm, rotary position embeddings (RoPE), and SwiGLU activations at two scales (\Cref{tab:arch}).

\begin{table}[h]
\centering
\caption{Model configurations.}\label{tab:arch}
\begin{tabular}{lcc}
\toprule
 & \textbf{Small} & \textbf{Large} \\
\midrule
Parameters & 28\,M & 120\,M \\
Layers & 12 & 16 \\
Hidden dim & 384 & 768 \\
Attention heads & 6 & 12 \\
FFN dim (SwiGLU) & 1536 & 3072 \\
\bottomrule
\end{tabular}
\end{table}

The input is a token sequence:
\[
  \texttt{[BOS]}\;\texttt{[time\_control]}\;\texttt{[elo]}\;\texttt{[color]}\;\text{move}_1\;\text{move}_2\;\cdots\;\text{move}_t
\]
where each move is in UCI notation (\eg, \texttt{e2e4}, \texttt{g1f3}).
The vocabulary contains ${\sim}1{,}900$ tokens.
The model has no access to any board representation; piece positions must be inferred entirely from the move history.
At inference, the model generates moves via temperature-controlled sampling with no search.

\subsection{Training details}\label{sec:training_details}

All models are trained for 2 epochs with AdamW ($\beta_1{=}0.9$, $\beta_2{=}0.999$, weight decay $0.1$), learning rate $10^{-4}$ with cosine decay to $10^{-5}$, batch size 2048, warmup of 4000 steps, gradient clipping at 1.0, dropout $0.1$, and bf16 mixed precision.
Maximum sequence length is 170 tokens (P99.5 of the data distribution).
The 28\,M models train in ${\sim}2.5$ hours on a single NVIDIA H100; the 120\,M models train in ${\sim}70$ hours.
All models use Flash Attention 2 and \texttt{torch.compile} for throughput optimization (${\sim}134{,}000$ tokens/s on H100).

\subsection{Training data}\label{sec:data}

All data is sourced from the Lichess open database, spanning the full rating range (${\sim}600$--$2800$ Elo) in Bullet and Blitz time controls.
Each game produces two training sequences (one per color).
We additionally include Lichess puzzle data (1--5.8\,M puzzles) as supplementary sequences.
No Elo-based filtering is applied by default.
\Cref{tab:models} provides a complete overview of all model variants discussed in this paper.

\begin{table}[t]
\centering
\caption{Overview of all model variants. Entries marked with $\dagger$ use Elo-filtered data rather than all-Elo data.}\label{tab:models}
\begin{tabular}{lccccl}
\toprule
\textbf{Model} & \textbf{Params} & \textbf{Data} & \textbf{Weighting} & \textbf{Illegal} & \textbf{Role in paper} \\
\midrule
V1.0   & 28\,M  & 10\,M all-Elo  & Uniform    & 1.06\% & Baseline \\
V1.0w  & 28\,M  & 10\,M all-Elo  & Linear     & 1.13\% & Weighting ablation \\
V1.1$\dagger$ & 28\,M  & high-Elo only  & Uniform    & 1.82\% & Filtering experiment \\
V2.0   & 120\,M & 50\,M all-Elo  & Uniform    & 0.22\% & Scaling ablation \\
V2.2   & 120\,M & 50\,M all-Elo  & Linear     & 0.26\% & \textbf{Final model} (Lichess 2570) \\
V2.3   & 120\,M & 100\,M all-Elo & Exponential& 0\%    & Over-weighting \\
\bottomrule
\end{tabular}
\end{table}

\begin{table}[h]
\centering
\caption{Training data scales.}\label{tab:data}
\begin{tabular}{lccc}
\toprule
\textbf{Dataset} & \textbf{Games} & \textbf{Sequences} & \textbf{Tokens} \\
\midrule
Scale 1 (V1 series) & ${\sim}10$\,M & ${\sim}20$\,M & ${\sim}1.4$\,B \\
Scale 2 (V2.2)      & ${\sim}50$\,M & ${\sim}100$\,M & ${\sim}7$\,B \\
Scale 3 (V2.3)      & ${\sim}100$\,M & ${\sim}150$\,M & ${\sim}14$\,B \\
\bottomrule
\end{tabular}
\end{table}

\subsection{Elo-weighted training}\label{sec:weighting_method}

Rather than selecting which games to include, we modulate gradient contributions via sample-level loss weighting:
\begin{equation}\label{eq:weighted_loss}
  \mathcal{L} = \sum_i w(e_i)\,\ell(x_i),
\end{equation}
where $\ell(x_i)$ is the cross-entropy loss for sequence $i$ and $w(e_i)$ depends on the average Elo of the two players.
We define the \emph{weighting intensity} $r = w(e_{\max})/w(e_{\min})$ and investigate three schemes (\Cref{tab:weighting}).

\begin{table}[h]
\centering
\caption{Weighting schemes.}\label{tab:weighting}
\begin{tabular}{lccc}
\toprule
\textbf{Scheme} & \textbf{Function} & \textbf{$r$} & \textbf{High-Elo gradient} \\
\midrule
Uniform      & $w(e)=1$ & $1{:}1$ & ${\sim}5\%$ \\
Linear       & $w(e)=\mathrm{clip}(\alpha(e - e_{\min}),\, w_{\min},\, 1)$ & ${\sim}20{:}1$ & ${\sim}45\%$ \\
Exponential  & $w(e)=\exp(\beta(e - e_{\mathrm{ref}}))$ & ${\sim}200{:}1$ & ${\sim}85\%$ \\
\bottomrule
\end{tabular}
\end{table}

\subsection{Analysis framework: tracking and decision quality}\label{sec:framework}

A sequence model predicting chess moves must implicitly solve two subproblems:

\paragraph{State tracking ($T$).}
Reconstructing the board from the move sequence.
We measure $T$ primarily via the \emph{illegal-move rate}: each illegal move is a direct, unambiguous signal of tracking failure.
We additionally use linear probes \citep{li2023emergent,karvonen2024emergent} on hidden states to measure internal board-representation accuracy.

\paragraph{Decision quality ($Q$).}
Selecting human-like moves given an (approximately) accurate board representation.
We measure $Q$ through CPL, move-prediction accuracy on high-rated subsets, and Stockfish agreement rate.

These jointly determine playing strength.
A model with perfect $T$ but poor $Q$ plays legal but weak moves.
A model with strong $Q$ but degraded $T$ periodically blunders material---and a single piece blunder typically decides the game.
We formalize this as the bottleneck in \Cref{eq:bottleneck}: $P \le \min(T, Q)$.

\FloatBarrier
\section{The Diversity--Quality Paradox}\label{sec:paradox}

The natural approach to improving playing strength is to filter out low-rated games and train only on high-quality data.
We show that this intuition is not merely wrong but \emph{predictably} wrong: filtering consistently degrades performance, and the mechanism reveals the dual-capability structure.

\subsection{Why tracking and decision quality require opposing data}\label{sec:opposing}

We define the \emph{effective diversity} of dataset $D$ under weighting $w$ as the number of distinct board states receiving at least $\theta$ units of weighted exposure:
\begin{equation}
  \mathrm{Div}(D) = \bigl|\{s : s \text{ appears in } D \text{ at least } \theta \text{ times}\}\bigr|.
\end{equation}
The \emph{effective quality} is the weighted average move quality:
\begin{equation}
  \mathrm{Qual}(D, w) = \frac{\sum_e w(e)\,n_e\,q(e)}{\sum_e w(e)\,n_e},
\end{equation}
where $q(e)$ is the average move quality (inverse CPL) at rating $e$.

Low-rated data contributes high diversity (unusual positions from unconventional play) but low quality.
High-rated data contributes high quality but narrow diversity (standard openings, typical structures).
Filtering removes the former: $\mathrm{Qual}$ rises, $\mathrm{Div}$ crashes, and under $P \le \min(T, Q)$ the bottleneck switches from $Q$ to $T$ (\Cref{tab:filtering_theory}).

\begin{table}[h]
\centering
\caption{Predicted effect of filtering under the dual-capability bottleneck.}\label{tab:filtering_theory}
\begin{tabular}{lccccc}
\toprule
 & Div & Qual & $T$ & $Q$ & $P = \min(T,Q)$ \\
\midrule
All Elo (before)       & High & Low  & High & Low  & Limited by $Q$ \\
High Elo only (after)  & $\downarrow\!\downarrow$ & $\uparrow$ & $\downarrow\!\downarrow$ & $\uparrow$ & Limited by $T$ $\to$ worse \\
\bottomrule
\end{tabular}
\end{table}

\subsection{Experimental validation}\label{sec:filtering_exp}

We compare two 28\,M models trained on datasets of comparable size:

\begin{itemize}[nosep]
  \item \textbf{V1.0} (baseline): 9.6\,M Lichess games spanning all Elo levels (${\sim}20$\,M sequences, ${\sim}1.4$\,B tokens, 2 epochs).
  \item \textbf{V1.1} (filtered): retaining only Bullet 2500+ and Blitz 2200+ games, supplemented with tracking patches and puzzles (${\sim}25.8$\,M sequences, 2 epochs).
\end{itemize}

\begin{table}[h]
\centering
\caption{Filtering experiment: V1.0 (all Elo) vs.\ V1.1 (filtered).}\label{tab:filtering}
\begin{tabular}{lccc}
\toprule
\textbf{Metric} & \textbf{V1.0} & \textbf{V1.1} & \textbf{Interpretation} \\
\midrule
Illegal-move rate       & 1.06\% & 1.82\% (+72\%)  & $T$ degraded sharply \\
Opening Top-1           & 50.0\% & 53.8\% (+3.8\,pp) & $Q$ improved marginally \\
Middlegame Top-1        & 43.5\% & 39.9\% ($-3.6$\,pp) & Middlegame collapsed \\
Endgame Top-1           & 46.8\% & 40.4\% ($-6.4$\,pp) & Endgame collapsed \\
Head-to-head (100 games) & --- & V1.0 wins & $P$ decreased \\
\bottomrule
\end{tabular}
\end{table}

The results match the theoretical prediction point by point.
The asymmetry of damage is striking: a 3.8\,pp improvement in opening accuracy is vastly outweighed by 3.6--6.4\,pp degradation in middlegame and endgame---precisely the phases where tracking is most challenged.

Cross-evaluation on V1.0's validation set confirms the result is not an artifact of distribution shift: validation loss worsens from 1.75 to 2.01, Top-1 accuracy drops from 46.5\% to 41.3\%, and illegal rate rises from 1.06\% to 2.49\%.

Scaling alone does not rescue filtering.
A 58\,M model trained on pure 2000+ data performs worse than V1.0 on both Top-1 accuracy (40.3\% vs.\ 46.5\%) and illegal rate (2.68\% vs.\ 1.06\%), losing the direct match 0--9--11.

\subsection{Direct evidence from internal representations}\label{sec:probes}

To move beyond behavioral metrics, we train linear board probes on the hidden states of V1.0 and V1.1.
The probe dataset contains 315,606 positions from natural bullet decision points (252,967 train / 31,965 validation / 30,674 test), with 278,760 non-standard and 36,846 standard positions.
For each layer, an independent linear probe predicts one of 13 labels (empty, six white piece types, six black piece types) for each of the 64 squares.

\begin{figure}[htbp]
\centering
\includegraphics[width=\columnwidth]{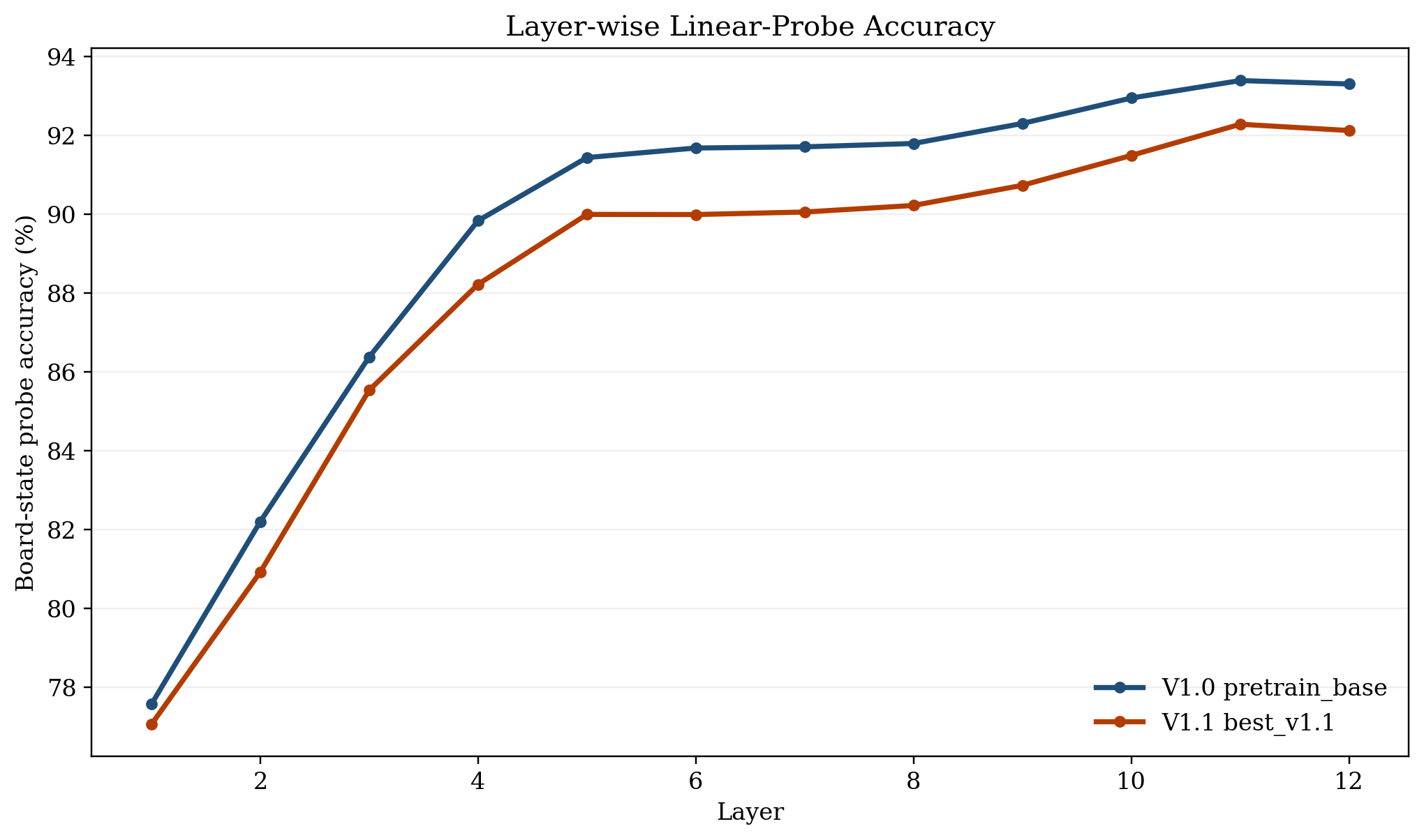}
\caption{Layer-wise board-state probe accuracy. The all-Elo base model (V1.0) maintains higher square accuracy throughout the network, with both models peaking in the final layer.}
\label{fig:probes}
\end{figure}

\begin{figure}[htbp]
\centering
\includegraphics[width=\columnwidth]{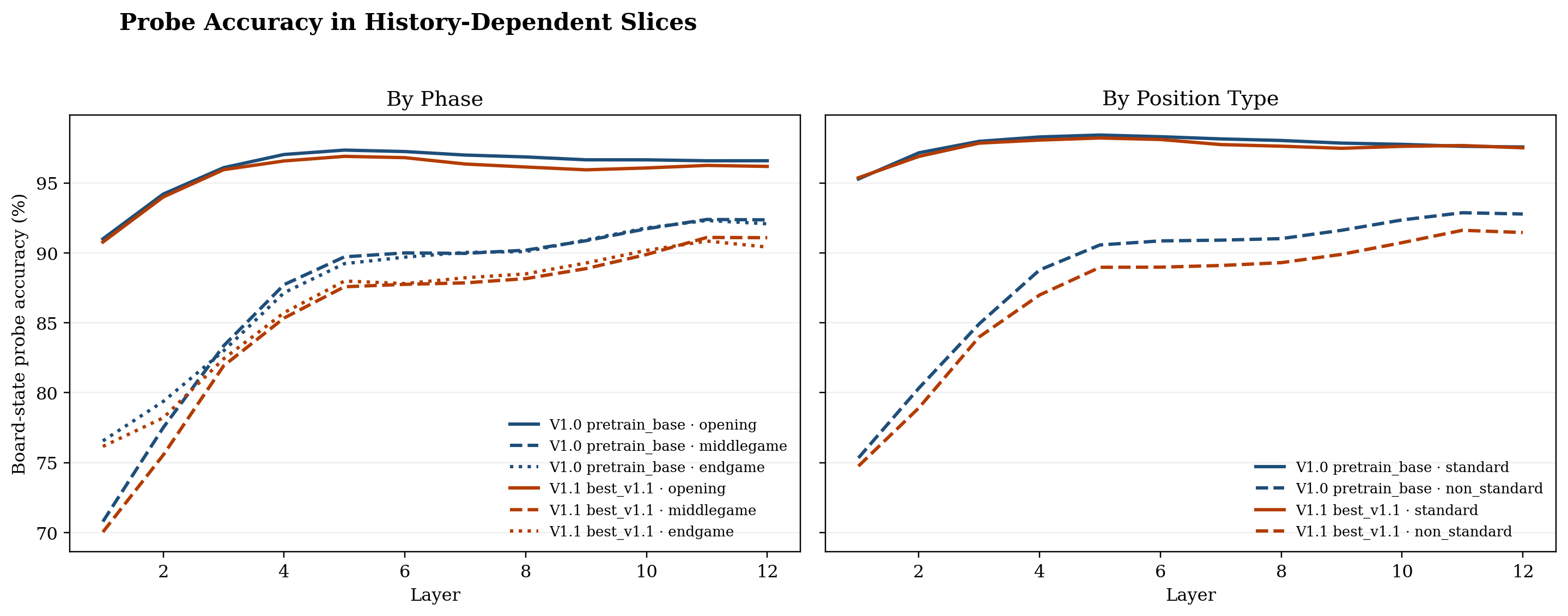}
\caption{Probe accuracy by evaluation slice. The gap between V1.0 and V1.1 is smallest on standard and opening positions, and largest on non-standard, middlegame, and endgame slices---precisely where low-Elo games contribute the most diverse positions.}
\label{fig:probes_slices}
\end{figure}

Results (\Cref{fig:probes,fig:probes_slices}):
\begin{itemize}[nosep]
  \item Best overall square accuracy: 93.40\% (V1.0) vs.\ 92.29\% (V1.1).
  \item On \emph{standard} positions: nearly identical (97.57\% vs.\ 97.51\%).
  \item On \emph{non-standard} positions: substantial gap (92.78\% vs.\ 91.45\%).
  \item The gap widens further in middlegame (92.36\% vs.\ 91.08\%) and endgame (92.08\% vs.\ 90.41\%).
\end{itemize}

The probes upgrade the filtering argument from behavioral correlation to \emph{mechanistic evidence}: removing low-Elo data degrades the internal board representation precisely where broad positional coverage matters most.

\subsection{Summary}

Under $P \le \min(T, Q)$, any strategy that sacrifices positional diversity to improve move quality will shift the bottleneck from $Q$ to $T$.
Filtering is the most extreme form: it hard-deletes samples, causing irreversible diversity loss.
The problem is not the \emph{degree} of filtering but the \emph{kind} of operation.
This motivates the question: can we reduce the influence of low-quality moves on decision learning without removing the positions they generate from tracking training?

\FloatBarrier
\section{The Weighting Sweet Spot}\label{sec:sweetspot}

\subsection{Weighting versus filtering: a qualitative difference}\label{sec:weight_vs_filter}

Filtering removes low-rated samples entirely: $D_{\mathrm{filtered}} = \{(x,e) \in D : e > e_{\mathrm{thresh}}\}$.
Every position exclusive to low-rated games contributes exactly zero to tracking training.

Weighting retains all samples but modulates gradient contributions:
$\nabla\mathcal{L} = \sum_i w(e_i)\,\nabla\ell(x_i)$.
Under linear weighting with $r{=}20$, low-Elo data contributes ${\sim}5\%$ of the total gradient, compared to ${\sim}70\%$ under uniform weighting and 0\% under filtering.

The critical insight is that weighting affects \emph{both} capabilities simultaneously:
\begin{itemize}[nosep]
  \item $Q$ increases with $r$: the gradient becomes dominated by high-rated samples.
  \item $T$ decreases with $r$: gradient contribution from diverse low-rated positions diminishes.
\end{itemize}
Weighting is therefore not monotonically beneficial.
There must exist an intermediate $r$ where the two capabilities are balanced.

\subsection{Theoretical formulation of the sweet spot}\label{sec:sweetspot_theory}

Let $r$ denote the weighting intensity.
In log-space, we approximate each capability's dependence on $r$ as linear:
\begin{align}
  T(r) &= T_0 - \alpha \ln r, \label{eq:T}\\
  Q(r) &= Q_0 + \beta \ln r, \label{eq:Q}
\end{align}
where $T_0, Q_0$ are baselines under uniform weighting ($r{=}1$), and $\alpha, \beta$ capture sensitivity to weighting.
The empirical finding $T_0 > Q_0$ reflects that uniform weighting yields serviceable tracking but weak decisions.

Since $P = \min(T(r), Q(r))$, the optimum is at the crossover:
\begin{equation}\label{eq:rstar}
  r^{*} = \exp\!\biggl(\frac{T_0 - Q_0}{\alpha + \beta}\biggr).
\end{equation}

\paragraph{Dependence on model capacity.}
Both $T_0$ and $\alpha$ depend on model size $C$.
Larger models have higher $T_0(C)$ (better tracking baseline) and lower $\alpha(C)$ (less sensitive to reduced tracking signal), shifting the sweet spot rightward: $r^{*}(C)$ increases with $C$.
In other words, larger models tolerate more aggressive weighting.

\subsection{Empirical validation}\label{sec:weight_exp}

We distinguish three regimes across the available experiments.

\paragraph{Unweighted regime ($r{=}1$).}
Both V1.0 (28\,M) and V2.0 (120\,M) use uniform weighting.
V1.0 reaches ${\sim}2000$ Lichess bullet with 1.06\% illegal rate; V2.0 improves to 0.22\% illegal rate and 50.85\% Top-1, but gains only modestly in head-to-head play (80--54 over 200 games vs.\ V1.0).
The consistent pattern across both scales is that decision quality, rather than tracking, limits practical performance under uniform weighting.

\paragraph{At the sweet spot ($r{\approx}20$, V2.2, 120\,M).}
Linear Elo weighting with 20:1 ratio on 50\,M all-Elo games (\Cref{tab:v22}).

\begin{table}[h]
\centering
\caption{V2.2 performance (120\,M, linear weighting, $r{=}20$).}\label{tab:v22}
\begin{tabular}{lc}
\toprule
\textbf{Metric} & \textbf{Value} \\
\midrule
Illegal-move rate & 0.26\% \\
Top-1 accuracy & 51.2\% \\
Lichess bullet Elo & 2570 (253 games) \\
vs.\ V1.0 (100 games) & 66W--8L--26D \\
\bottomrule
\end{tabular}
\end{table}

Both $T$ and $Q$ reach high levels simultaneously.
Diagnosis: $T \approx Q$---we are at or near the sweet spot.

\paragraph{Right of the sweet spot ($r{\approx}200$, V2.3, 120\,M).}
Exponential weighting with 200:1 ratio on 100\,M games (\Cref{tab:v23}).

\begin{table}[h]
\centering
\caption{V2.2 (linear, $r{\approx}20$) vs.\ V2.3 (exponential, $r{\approx}200$).}\label{tab:v23}
\begin{tabular}{lcc}
\toprule
\textbf{Metric} & \textbf{V2.2} & \textbf{V2.3} \\
\midrule
Validation loss & 1.48 & ${\sim}1.40$ (lower) \\
Practical play  & Lichess 2570 & Weaker \\
Elo conditioning (2600 vs.\ 1200) & Works (27:13) & Broken (17:16 $\approx$ random) \\
Endgame quality & Normal & Collapsed \\
\bottomrule
\end{tabular}
\end{table}

Despite achieving lower validation loss, V2.3 plays worse.
The exponential weighting reduced low-Elo gradients to ${\sim}0.5\%$---functionally zero---recreating the filtering problem.
$T$ re-emerged as the bottleneck.

\subsection{Full factorial ablation: isolating weighting from scaling}\label{sec:ablation}

To disentangle the contributions of scaling and weighting, we complete the full $2 \times 2$ factorial design by training V2.0 (120\,M, uniform weighting) and V1.0w (28\,M, linear weighting), in addition to V1.0 (28\,M, uniform) and V2.2 (120\,M, linear).
\Cref{tab:factorial} summarizes the offline metrics; \Cref{tab:h2h} reports all pairwise head-to-head results.

\begin{table}[h]
\centering
\caption{Full $2 \times 2$ factorial: offline metrics. Weighting slightly increases illegal rate (degraded tracking) while improving Top-1 accuracy (better decisions) at both scales.}\label{tab:factorial}
\begin{tabular}{lcccc}
\toprule
 & \textbf{V1.0} & \textbf{V1.0w} & \textbf{V2.0} & \textbf{V2.2} \\
 & 28M, $r{=}1$ & 28M, $r{=}20$ & 120M, $r{=}1$ & 120M, $r{=}20$ \\
\midrule
Illegal rate & 1.06\% & 1.13\% & 0.22\% & 0.26\% \\
Top-1 accuracy & 46.5\% & 47.0\% & 50.85\% & 51.2\% \\
\bottomrule
\end{tabular}
\end{table}

\begin{table}[h]
\centering
\caption{Pairwise head-to-head results. $p$-values from two-sided binomial tests on decisive games (H$_0$: equal strength).}\label{tab:h2h}
\begin{tabular}{lcccc}
\toprule
\textbf{Matchup} & \textbf{W--L--D} & \textbf{Games} & \textbf{Score} & \textbf{$p$} \\
\midrule
V1.0w vs.\ V1.0 (weighting, 28M)    & 40--23--37  & 100 & 58.5\% & 0.043 \\
V2.0 vs.\ V1.0 (scaling, $r{=}1$)   & 80--54--66  & 200 & 56.5\% & 0.030 \\
V2.0 vs.\ V2.2 (weighting, 120M)    & 24--41--35  & 100 & 41.5\% & 0.046 \\
V2.2 vs.\ V1.0 (combined)           & 66--8--26   & 100 & 79.0\% & $<$0.001 \\
V2.2 vs.\ V1.0w (scaling, $r{=}20$) & 66--12--22  & 100 & 77.0\% & $<$0.001 \\
\bottomrule
\end{tabular}
\end{table}

Three findings emerge from the completed factorial:

\paragraph{Scaling fixes tracking.}
Moving from 28\,M to 120\,M at fixed uniform weighting (V1.0~$\to$~V2.0) drops the illegal rate from 1.06\% to 0.22\% ($5{\times}$ reduction) and raises Top-1 accuracy from 46.5\% to 50.85\%.
These are primarily tracking and representation gains: the larger model reconstructs the board far more reliably.
In head-to-head play, V2.0 beats V1.0 by 80--54 over 200 games ($p{=}0.030$)---significant but modest, consistent with scaling improving tracking without changing decision strategy.

\paragraph{Weighting trades tracking for decisions.}
At both scales, weighting \emph{slightly increases} the illegal rate (1.06\%~$\to$~1.13\% at 28\,M; 0.22\%~$\to$~0.26\% at 120\,M) while \emph{improving} Top-1 accuracy (46.5\%~$\to$~47.0\%; 50.85\%~$\to$~51.2\%).
This is exactly the tradeoff predicted by the $\min(T, Q)$ framework: upweighting high-Elo gradients improves decision quality at the cost of reduced tracking diversity.
The tradeoff is net-positive in head-to-head play: V1.0w beats V1.0 (40--23, $p{=}0.043$), and V2.2 beats V2.0 (41--24, $p{=}0.046$), confirming that the decision gains outweigh the small tracking cost.

\paragraph{The two interventions are complementary.}
Scaling primarily raises the tracking floor (illegal rate $5{\times}$ lower); weighting primarily raises the decision ceiling (Top-1 and head-to-head strength), at a small tracking cost that the larger model can absorb.
Neither alone achieves the full effect: V2.0 beats V1.0 modestly (80--54 over 200 games), V1.0w beats V1.0 similarly (40--23), but V2.2 dominates V1.0 decisively (66--8, $p{<}0.001$).
The combination produces the strongest practical-play gains: scaling provides the tracking headroom that allows weighting to improve decisions without crossing the tracking failure threshold.

\subsection{Tracking evidence across ablation models}\label{sec:probe_ablation}

\begin{figure}[htbp]
\centering
\includegraphics[width=\columnwidth]{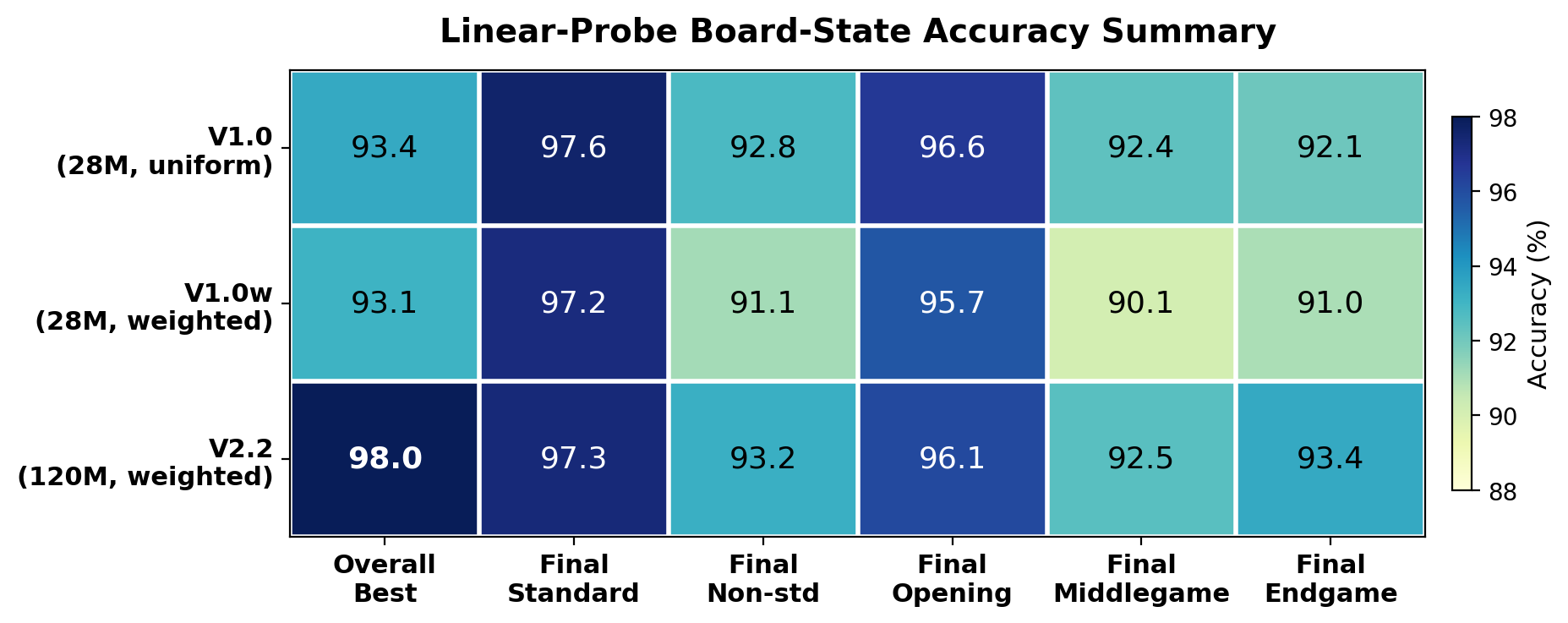}
\caption{Linear-probe board-state accuracy for three models (V1.0, V1.0w, V2.2) under unified evaluation. Scaling from 28\,M to 120\,M produces a large tracking improvement (93.4\% $\to$ 98.0\%), while weighting at fixed 28\,M scale does not improve---and slightly reduces---probe accuracy.}
\label{fig:three_probes}
\end{figure}

To provide mechanistic support for the factorial results, we probe three models from the ablation design (V1.0, V1.0w, V2.2) using linear board-state probes under a unified protocol (\Cref{fig:three_probes}).
The 120\,M model (V2.2) achieves 98.0\% best overall accuracy, a substantial jump from both 28\,M models (93.4\% for V1.0, 93.1\% for V1.0w).
This provides direct internal evidence that scaling is the primary driver of tracking improvement.
V1.0w's probe accuracy is \emph{slightly lower} than V1.0's (93.1\% vs.\ 93.4\%), consistent with the behavioral finding that weighting trades a small amount of tracking fidelity for improved decision quality.

\subsection{Validation loss is not performance}\label{sec:val_loss}

The most counterintuitive finding is that V2.3 achieves the \emph{lowest} validation loss \emph{and} zero illegal moves while playing the \emph{worst} among 120\,M variants.
Every standard offline metric---validation loss, illegal rate, Top-1 on high-Elo subsets---favors V2.3, yet it loses to V2.2 in practice.
This is a structural consequence of the bottleneck.

Validation loss averages over all positions, dominated by high-weighted positions where predictions are most confident.
Playing strength, by contrast, is determined by \emph{worst-case} tracking: a single piece blunder costs the game regardless of how well the model handles standard positions.
Validation loss measures the average; practical play is determined by the minimum.

This is a specific instance of Goodhart's Law applied to capability-structured systems: when performance is governed by a bottleneck over component capabilities, optimizing an aggregate metric can mask deterioration of the binding constraint.

\subsection{Temperature-dependent checkpoint ordering}\label{sec:temperature}

Checkpoint evaluation reveals a further disconnect between likelihood and strength.
In matches between checkpoints e1\_05 and e1\_09 (\Cref{tab:temp}), deterministic decoding yields a tie, low temperature favors the earlier checkpoint, and higher temperature favors the later checkpoint.

\begin{table}[h]
\centering
\caption{Checkpoint ordering depends on sampling temperature.}\label{tab:temp}
\begin{tabular}{lccc}
\toprule
\textbf{Temperature} & \textbf{e1\_09 wins} & \textbf{e1\_05 wins} & \textbf{Draws} \\
\midrule
0.0 & 50 & 50 & 0 \\
0.1 & 28 & 49 & 23 \\
0.2 & 31 & 43 & 26 \\
0.3 & 43 & 39 & 18 \\
\bottomrule
\end{tabular}
\end{table}

Additional training changes the tail of the move distribution more than the top-1 logit.
At higher temperatures, cleaner tail probabilities translate to stronger play.
Model selection is therefore not independent of inference policy: the best checkpoint is a function of the sampling temperature at which the model will play.

\FloatBarrier
\section{Intra-Game Degeneration and the Aligned Cliff}\label{sec:degeneration}

All sequence models we trained exhibit a recurring late-stage failure pattern in self-play.
However, this pattern is not a single global collapse boundary shared by all games.
Instead, individual games contain their own \emph{degeneration points}: local transitions after which severe errors become substantially more frequent.

\subsection{Degeneration points within games}\label{sec:degen_points}

We define a catastrophic blunder indicator for each move $t$, smooth with a sliding window, and identify the \emph{degeneration point} $t_{\mathrm{deg}}$ as the earliest sustained transition into a high-risk regime---not the first isolated mistake, but the onset of persistent instability.

\begin{itemize}[nosep]
  \item In the baseline model (pretrain\_base), 632/1000 games (63.2\%) exhibit a detectable degeneration point; in the final model (V2.2), only 195/1000 (19.5\%).
  \item Median $t_{\mathrm{deg}}$ shifts from move~26 (baseline) to move~31 (final model).
  \item Normalized by game length: median degeneration position shifts from 0.588 to 0.694.
\end{itemize}

\begin{figure}[htbp]
\centering
\includegraphics[width=\columnwidth]{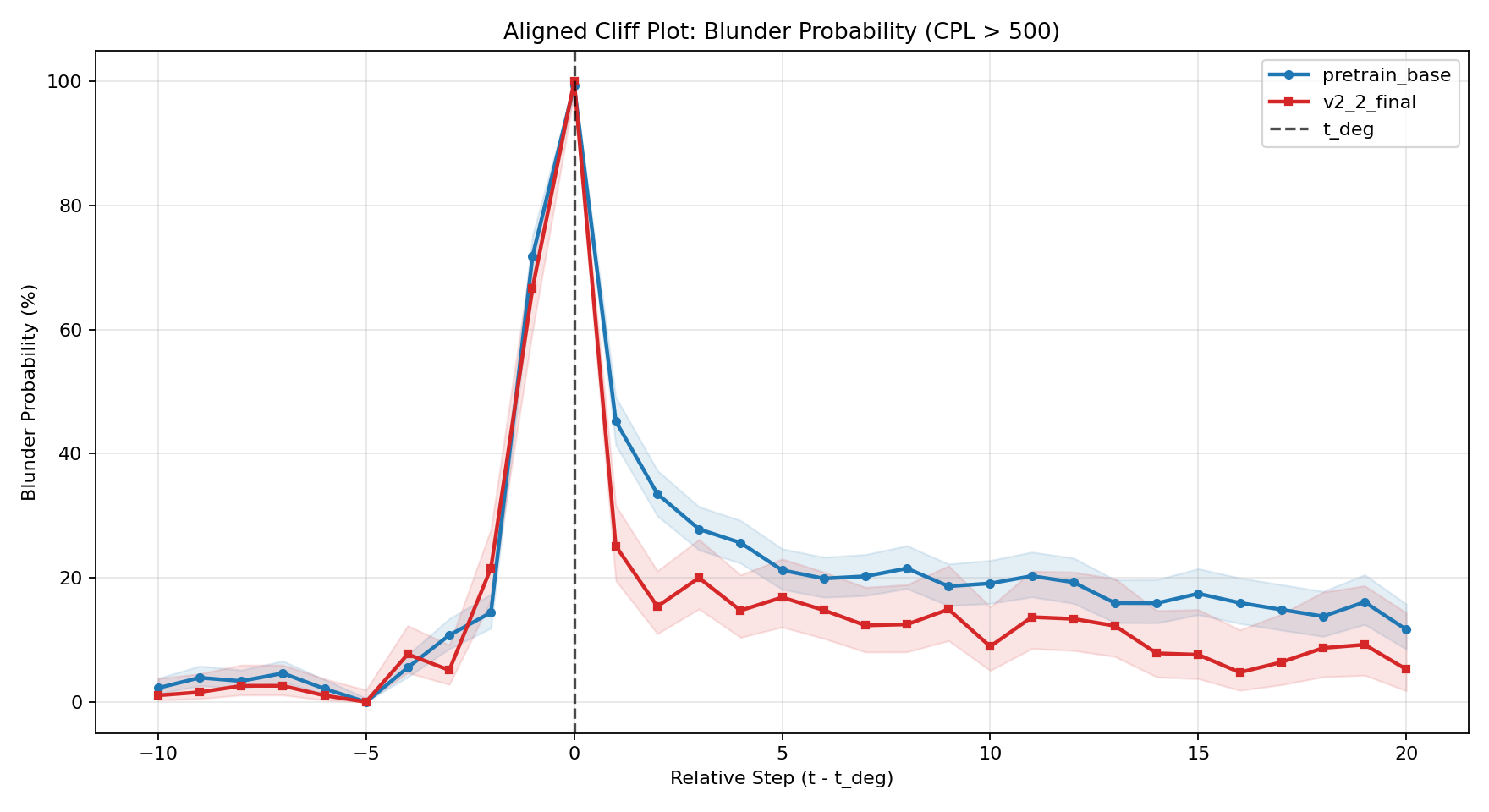}
\caption{Aligned cliff plot of catastrophic blunder probability (CPL~$>$~500) around $t_{\mathrm{deg}}$. Each game is shifted so that its degeneration point occurs at relative step~0. A sharp spike at $t_{\mathrm{deg}}$, followed by persistently elevated risk, reveals a cliff-like transition obscured in unaligned averages.}
\label{fig:cliff}
\end{figure}

When all games are \emph{aligned} by $t_{\mathrm{deg}}$, a striking cliff-like structure appears (\Cref{fig:cliff}): catastrophic blunder probability is low before $t_{\mathrm{deg}}$, spikes sharply at $t_{\mathrm{deg}}$, and remains elevated afterward.
This is not simply ``more error later''; games contain localized transition moments.

\begin{figure}[htbp]
\centering
\includegraphics[width=0.85\columnwidth]{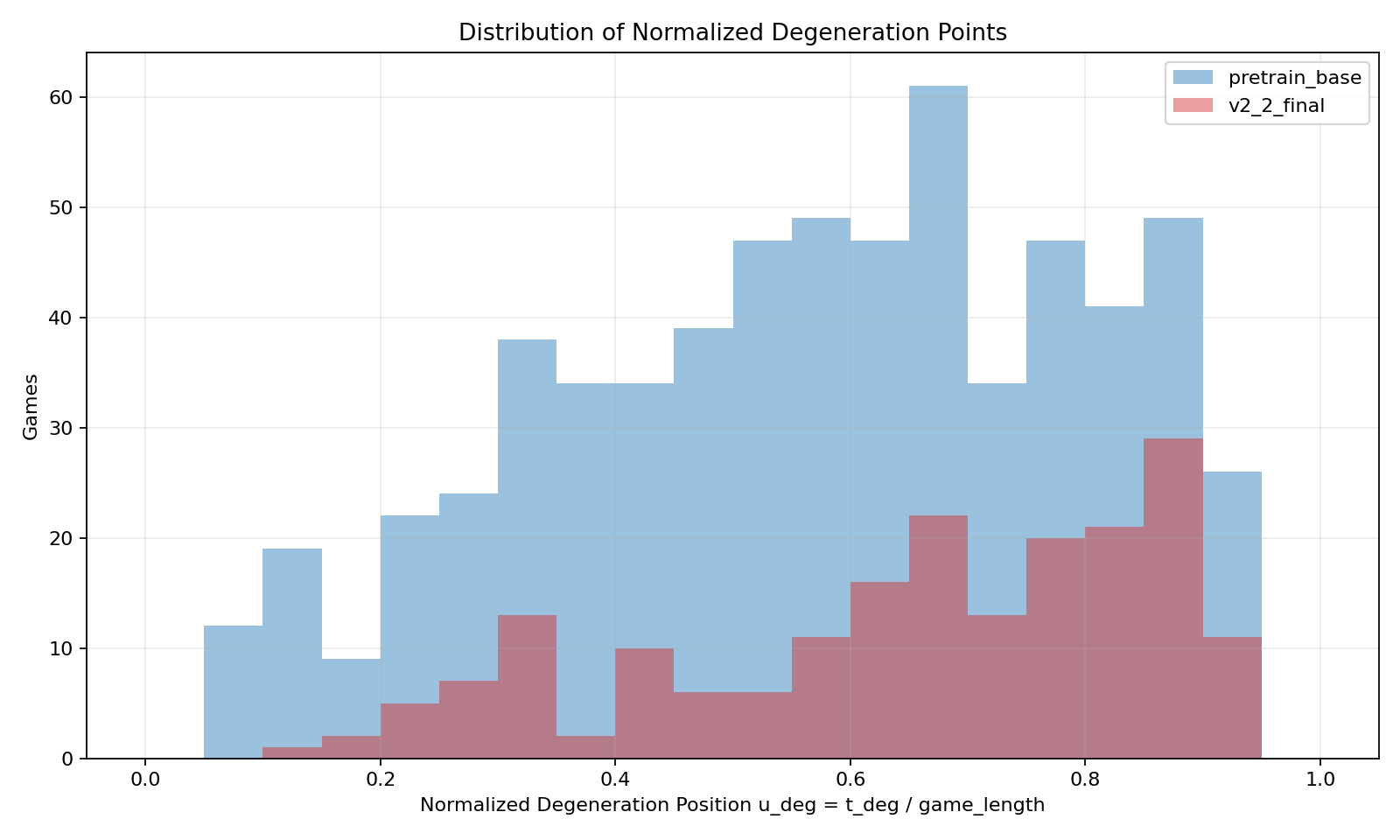}
\caption{Distribution of normalized degeneration positions $u_{\mathrm{deg}} = t_{\mathrm{deg}} / \text{game length}$. Degeneration occurs predominantly in later stages, and the final model shifts the distribution rightward.}
\label{fig:degen_dist}
\end{figure}

\begin{figure}[htbp]
\centering
\includegraphics[width=\columnwidth]{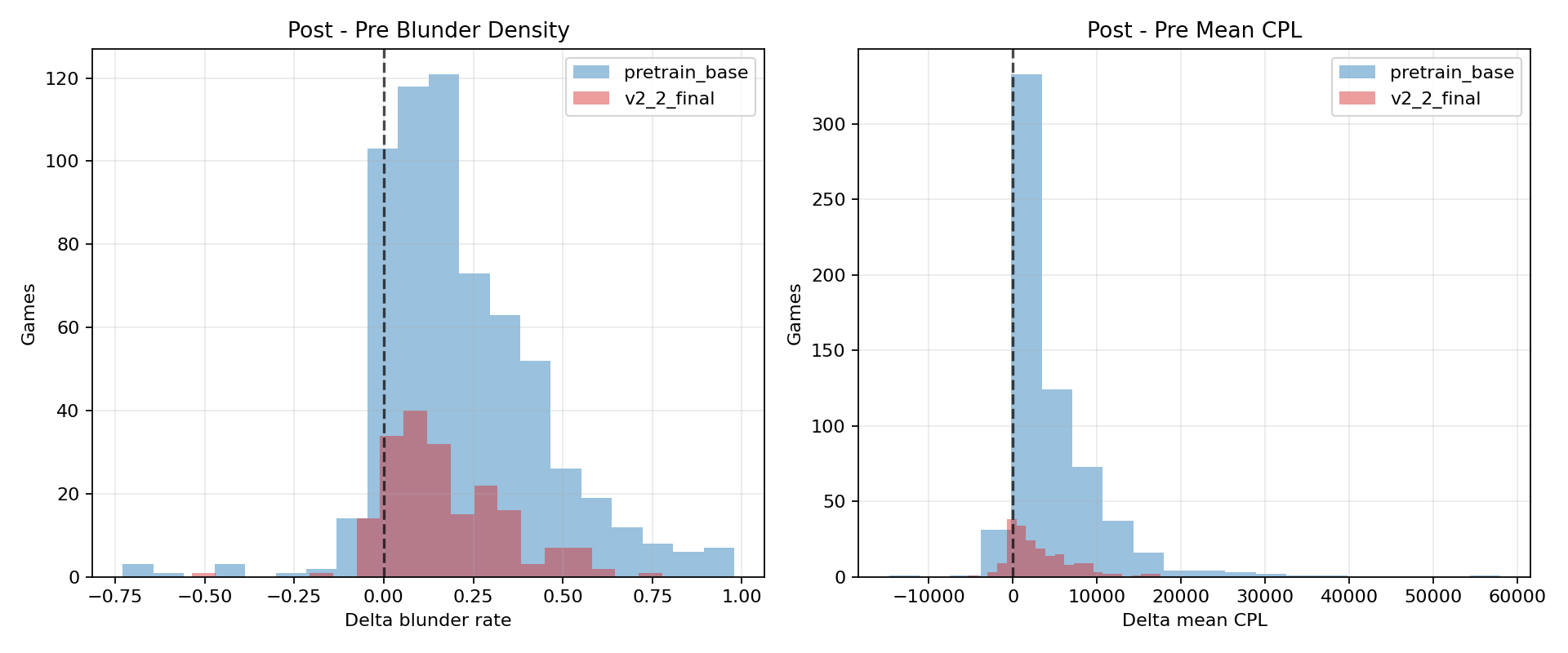}
\caption{Within-game error changes after the degeneration point. Left: change in blunder density. Right: change in mean CPL. For most games, both increase after $t_{\mathrm{deg}}$, validating it as a meaningful internal transition.}
\label{fig:pre_post}
\end{figure}

The final model exhibits the same qualitative structure but weakened: degeneration occurs in fewer games, later, and with lower post-degeneration risk.
Both blunder density and mean CPL are significantly higher after $t_{\mathrm{deg}}$ than before (Wilcoxon signed-rank $p < 10^{-25}$ for both models).

\subsection{A coverage-decay interpretation}\label{sec:coverage}

As self-play progresses, the set of reachable states expands while average training support declines.
We model this with a deliberately simple first-order approximation.
Let $S(t)$ denote the effective number of meaningfully distinct states reachable after depth $t$.
As a zeroth-order estimate, we assume exponential growth:
\begin{equation}
  S(t) \approx b^t,
\end{equation}
where $b$ is an \emph{effective} branching factor that accounts for transpositions and convergence among related continuations.
If the training set contains $N$ games, the average support per reachable state decays as $k(t) = N/b^t$.
Setting $k(t^{*}) = k_{\mathrm{crit}}$ (a minimum support threshold below which tracking becomes unreliable) yields a characteristic depth scale:
\begin{equation}\label{eq:tstar}
  t^{*} = \frac{\log(N/k_{\mathrm{crit}})}{\log b}.
\end{equation}

\paragraph{Scope and limitations of this formula.}
We emphasize that $t^{*}$ is a \emph{characteristic reliability horizon}, not a deterministic collapse boundary.
The formula rests on two simplifications that limit its precision:
\begin{itemize}[nosep]
  \item \textbf{Uniform branching.} In practice, the effective branching factor varies across game phases: openings are narrower (theory constrains choices), complex middlegames are wider, and endgames narrow again as material decreases. A phase-dependent $b(t)$ would yield a more accurate but less tractable expression.
  \item \textbf{Homogeneous support threshold.} The formula assumes a single $k_{\mathrm{crit}}$ for all positions, whereas in reality the model's tracking robustness depends on position complexity---simple endgames require less support than chaotic middlegames.
\end{itemize}
Despite these simplifications, the formula captures the qualitative structure well: it explains why degeneration points concentrate late in games (exponential state growth eventually exhausts any finite training set), why they form a \emph{distribution} rather than a single boundary (individual trajectories enter low-support regions at different depths), and why stronger models shift this distribution rightward (larger $N$ or better generalization effectively increases $N/k_{\mathrm{crit}}$).
We therefore use $t^{*}$ as an explanatory population-level scaffold, not as a per-game predictor.

\subsection{Why aligned cliffs emerge: compounding failure}\label{sec:compounding}

\begin{figure}[htbp]
\centering
\includegraphics[width=\columnwidth]{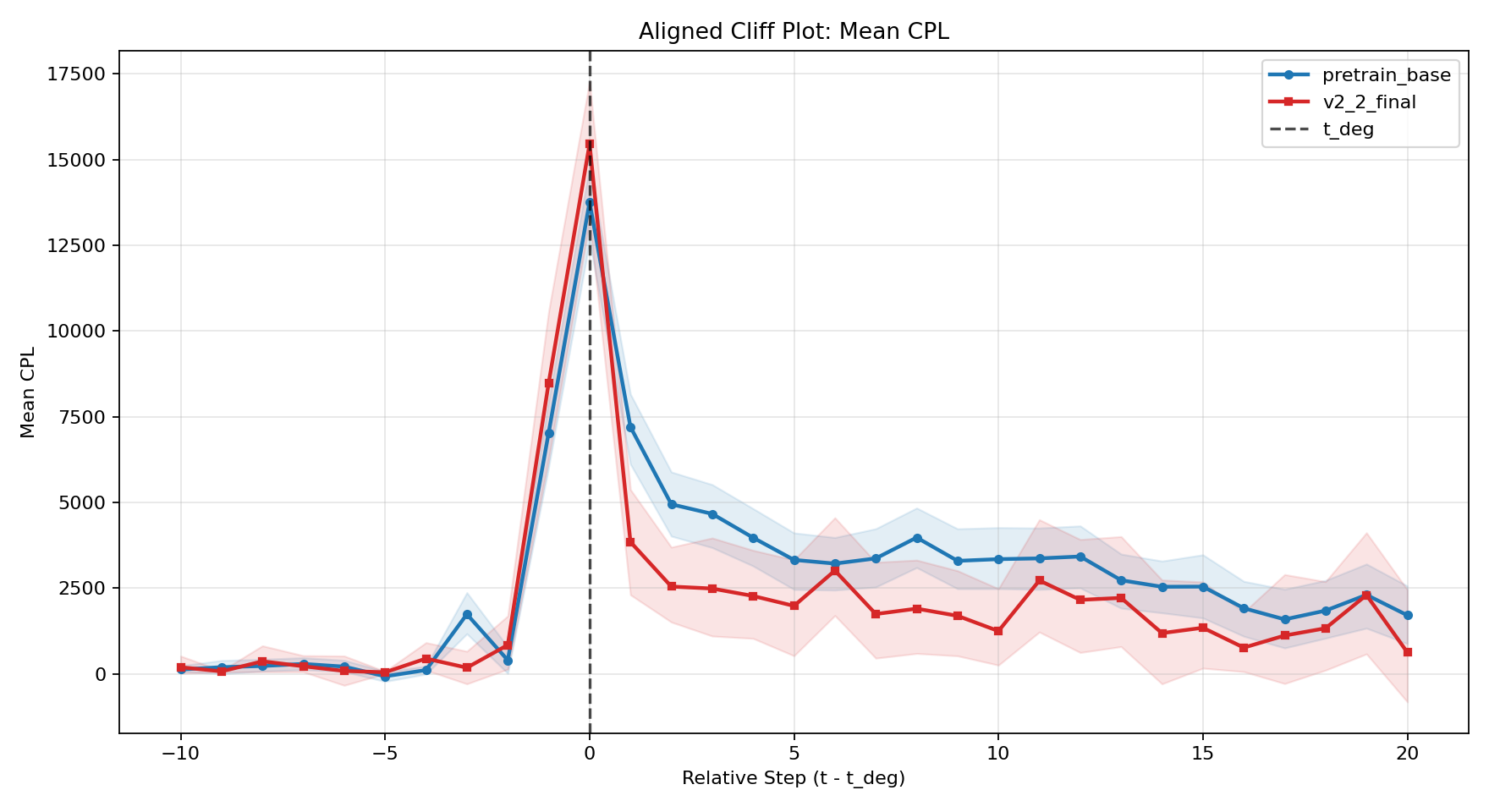}
\caption{Aligned cliff plot of mean CPL around $t_{\mathrm{deg}}$. Loss spikes at the degeneration point and remains substantially elevated, confirming a transition into a persistently worse error regime.}
\label{fig:cliff_cpl}
\end{figure}

\begin{figure}[htbp]
\centering
\includegraphics[width=\columnwidth]{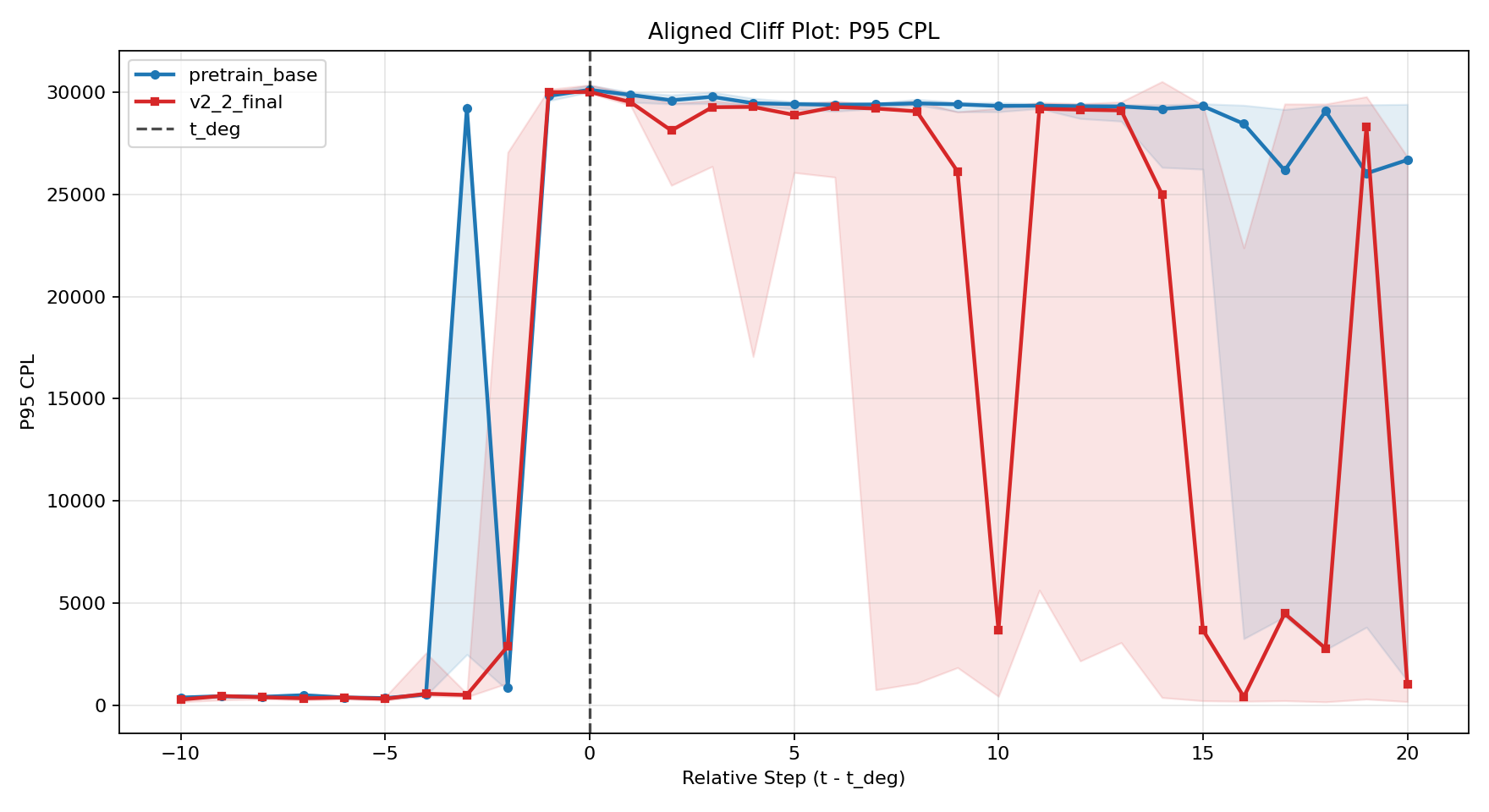}
\caption{Aligned P95 CPL around $t_{\mathrm{deg}}$. Tail losses rise sharply near degeneration, indicating that onset is dominated by extreme errors before the full average deterioration becomes visible.}
\label{fig:cliff_p95}
\end{figure}

Coverage decay explains why degeneration becomes more likely later, but not the cliff-like sharpness.
The cliff arises because errors \emph{compound}: a tracking error changes not only current move quality but the distribution of future states, pushing the game further off-distribution and increasing the probability of subsequent errors.
This produces a self-reinforcing transition into a persistently worse regime, visible in both mean CPL and P95 CPL aligned around $t_{\mathrm{deg}}$ (\Cref{fig:cliff_cpl,fig:cliff_p95}).

\subsection{Implications for scaling}\label{sec:degen_scaling}

Stronger models do not eliminate late-game fragility; they reduce the frequency of degeneration events, delay onset, and soften the post-degeneration regime.
The relevant evaluation question is not whether a model has a collapse boundary, but whether it can reduce degeneration frequency, delay onset, and limit severity.
On all three counts, the final model is substantially better.

\FloatBarrier
\section{Evaluation of Human-Likeness}\label{sec:humanlike}

We evaluate the model's human-likeness along three complementary axes: static move prediction against a strong baseline, a controlled test of history-dependent decision-making, and a deployment check on a public chess platform.

\subsection{Human move prediction}\label{sec:move_pred}

\begin{figure}[htbp]
\centering
\includegraphics[width=\columnwidth]{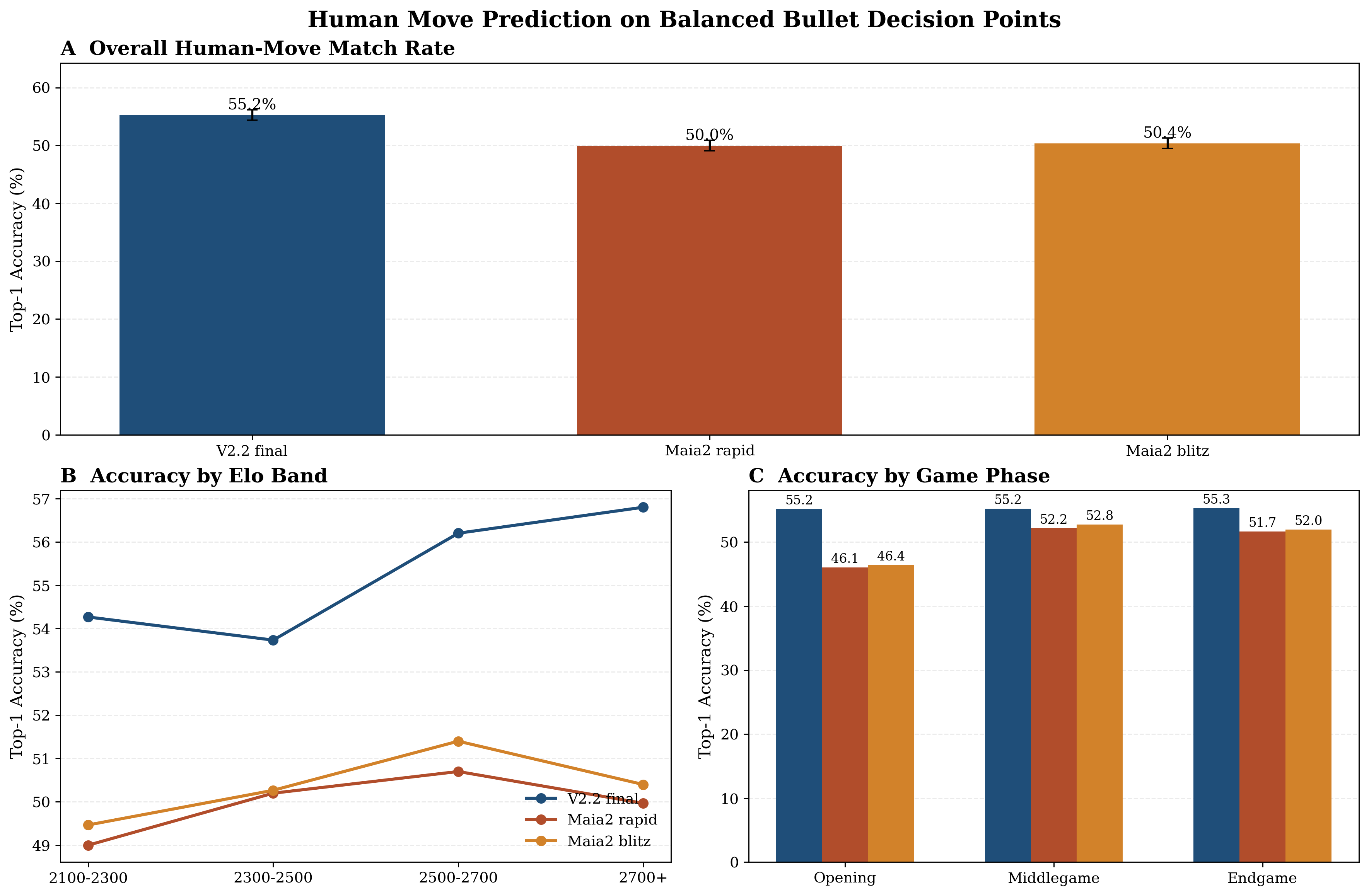}
\caption{Human move prediction comparison. Panel~A: overall Top-1 accuracy. Panel~B: accuracy by Elo band. Panel~C: accuracy by game phase. V2.2 outperforms both Maia-2 variants across all slices.}
\label{fig:maia2}
\end{figure}

We evaluate V2.2 against Maia-2 \citep{tang2024maia2}---the current state of the art for human move prediction---on 12,000 balanced bullet decision points sampled across four Elo bands (2100--2300, 2300--2500, 2500--2700, 2700+) and three game phases (opening, middlegame, endgame).
Both Maia-2 variants (rapid and blitz) are included.
All models are evaluated on the same held-out positions: for each position, V2.2 receives the full preceding move sequence, while Maia-2 receives the corresponding FEN---each model uses its native input format.
The evaluation set contains 12,000 decision points, balanced across four Elo bands and three game phases (1,000 positions per Elo~$\times$~phase cell).
The metric is Top-1 accuracy: whether the model's highest-probability move matches the human's actual move.
Error bars in \Cref{fig:maia2} denote 95\% binomial confidence intervals.

\begin{table}[htbp]
\centering
\caption{Human move prediction: Top-1 accuracy (\%) under native-input evaluation. V2.2 shows a consistent advantage across all slices; see text for fairness caveat.}\label{tab:maia2}
\small
\begin{tabular}{@{}lc@{\;\;}c@{\;\;}c@{\;\;}c@{\;\;}c@{\;\;}c@{\;\;}c@{\;\;}c@{}}
\toprule
\textbf{Model} & \textbf{Overall} & \textbf{2100--} & \textbf{2300--} & \textbf{2500--} & \textbf{2700+} & \textbf{Open.} & \textbf{Mid.} & \textbf{End.} \\
 & & \textbf{2300} & \textbf{2500} & \textbf{2700} & & & & \\
\midrule
V2.2 final   & \textbf{55.2} & \textbf{54.3} & \textbf{53.7} & \textbf{56.2} & \textbf{56.8} & \textbf{55.2} & \textbf{55.2} & \textbf{55.3} \\
Maia-2 rapid & 50.0 & 49.0 & 50.2 & 50.7 & 50.0 & 46.1 & 52.2 & 51.7 \\
Maia-2 blitz & 50.4 & 49.5 & 50.3 & 51.4 & 50.4 & 46.4 & 52.8 & 52.0 \\
\bottomrule
\end{tabular}
\end{table}

V2.2 achieves 55.2\% overall Top-1 accuracy, exceeding Maia-2 rapid (50.0\%) and Maia-2 blitz (50.4\%) by approximately 5 percentage points.
The advantage is consistent across all Elo bands and game phases (\Cref{tab:maia2,fig:maia2}).
Notably, V2.2's advantage is largest in the opening phase (+9.1\,pp vs.\ Maia-2 rapid, +8.8\,pp vs.\ Maia-2 blitz), where game history most strongly constrains the move distribution---precisely where FEN-only models lose information by discarding the preceding trajectory.

\paragraph{Fairness caveat.}
This comparison evaluates each model with its native input representation: V2.2 receives the full move sequence, while Maia-2 receives a FEN string.
The sequence input contains strictly more information (opening choice, move order, repetition history), so part of V2.2's advantage may stem from the \emph{representation} rather than the \emph{model}.
This is by design---a central claim of this paper is that sequence input enables capabilities that FEN input structurally cannot---but it means the comparison measures the combined effect of architecture, training, and input format, not model quality in isolation.
A controlled comparison would require training a Maia-2-scale model on move sequences, which we leave for future work.

\begin{figure}[htbp]
\centering
\includegraphics[width=\columnwidth]{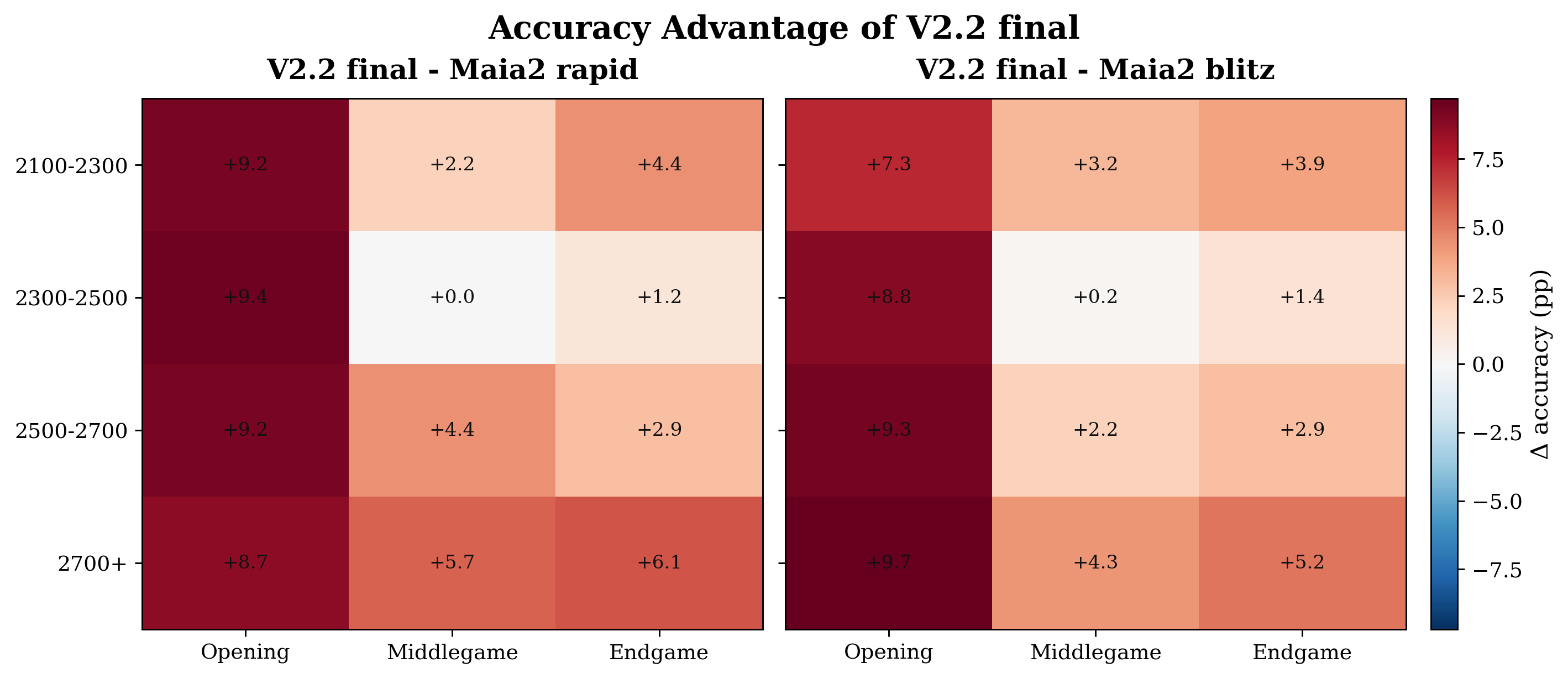}
\caption{Accuracy advantage of V2.2 over Maia-2 (percentage points) by Elo band and game phase. The advantage is largest in openings at lower Elo bands, where position diversity is highest.}
\label{fig:delta_heatmap}
\end{figure}

\subsection{Human-blunder alignment}\label{sec:blunder_align}

A human-like model should not only predict human moves---it should fail on the \emph{same} positions that humans find difficult.
We test this by measuring \emph{conditional blunder alignment}: when a human player makes a blunder (CPL~$>$~100), how much more likely is the model to also blunder on that position?

We evaluate on 314,418 positions from rated bullet games across four Elo bands.
For each position, we record whether the human and each model blundered, then compute the \emph{alignment lift}: the ratio of the model's blunder rate on human-blunder positions to its blunder rate on human-non-blunder positions.

\begin{table}[htbp]
\centering
\caption{Human-blunder alignment (CPL~$>$~100 threshold). V2.2 shows the highest conditional blunder probability and lift, indicating stronger alignment with human error patterns.}\label{tab:blunder}
\begin{tabular}{lccc}
\toprule
\textbf{Model} & \textbf{$P$(model blunder $|$ human blunder)} & \textbf{$P$(model blunder $|$ human $\neg$blunder)} & \textbf{Lift} \\
\midrule
V2.2 final   & \textbf{49.2\%} & 6.6\% & \textbf{7.4$\times$} \\
Maia-2 rapid & 43.3\% & 6.6\% & 6.5$\times$ \\
Maia-2 blitz & 44.8\% & 6.8\% & 6.6$\times$ \\
\bottomrule
\end{tabular}
\end{table}

\begin{figure}[htbp]
\centering
\includegraphics[width=\columnwidth]{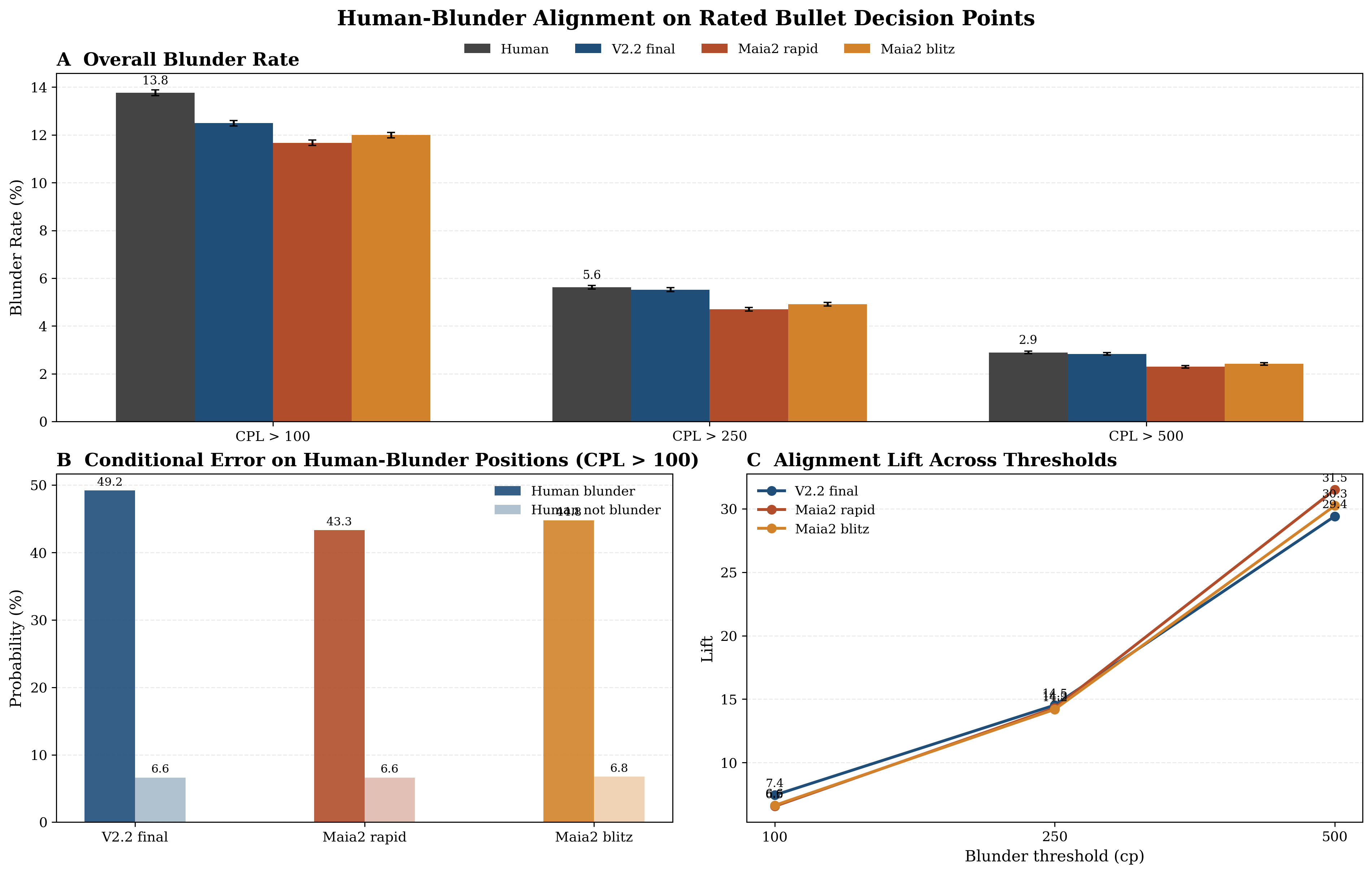}
\caption{Human-blunder alignment. Panel~A: overall blunder rates at three thresholds. Panel~B: conditional blunder probability on human-blunder vs.\ human-non-blunder positions. Panel~C: alignment lift increases with threshold severity, and V2.2 consistently leads.}
\label{fig:blunder_alignment}
\end{figure}

At CPL~$>$~100, V2.2's alignment lift (7.4$\times$) exceeds both Maia-2 variants (6.5--6.6$\times$), indicating that V2.2's errors are more concentrated on human-difficult positions.
We note that V2.2's overall blunder \emph{rate} is slightly higher than Maia-2's (12.5\% vs.\ 11.7--12.0\%), so the advantage is specifically in error \emph{overlap} with humans, not in making fewer mistakes.
At CPL~$>$~500 (catastrophic blunders), the picture is mixed: V2.2 maintains a higher conditional coincidence with human blunders (45.7\% vs.\ 38.4\%), but Maia-2 rapid achieves a slightly higher lift (31.5$\times$ vs.\ 29.4$\times$) due to its lower base rate.

\begin{figure}[htbp]
\centering
\includegraphics[width=\columnwidth]{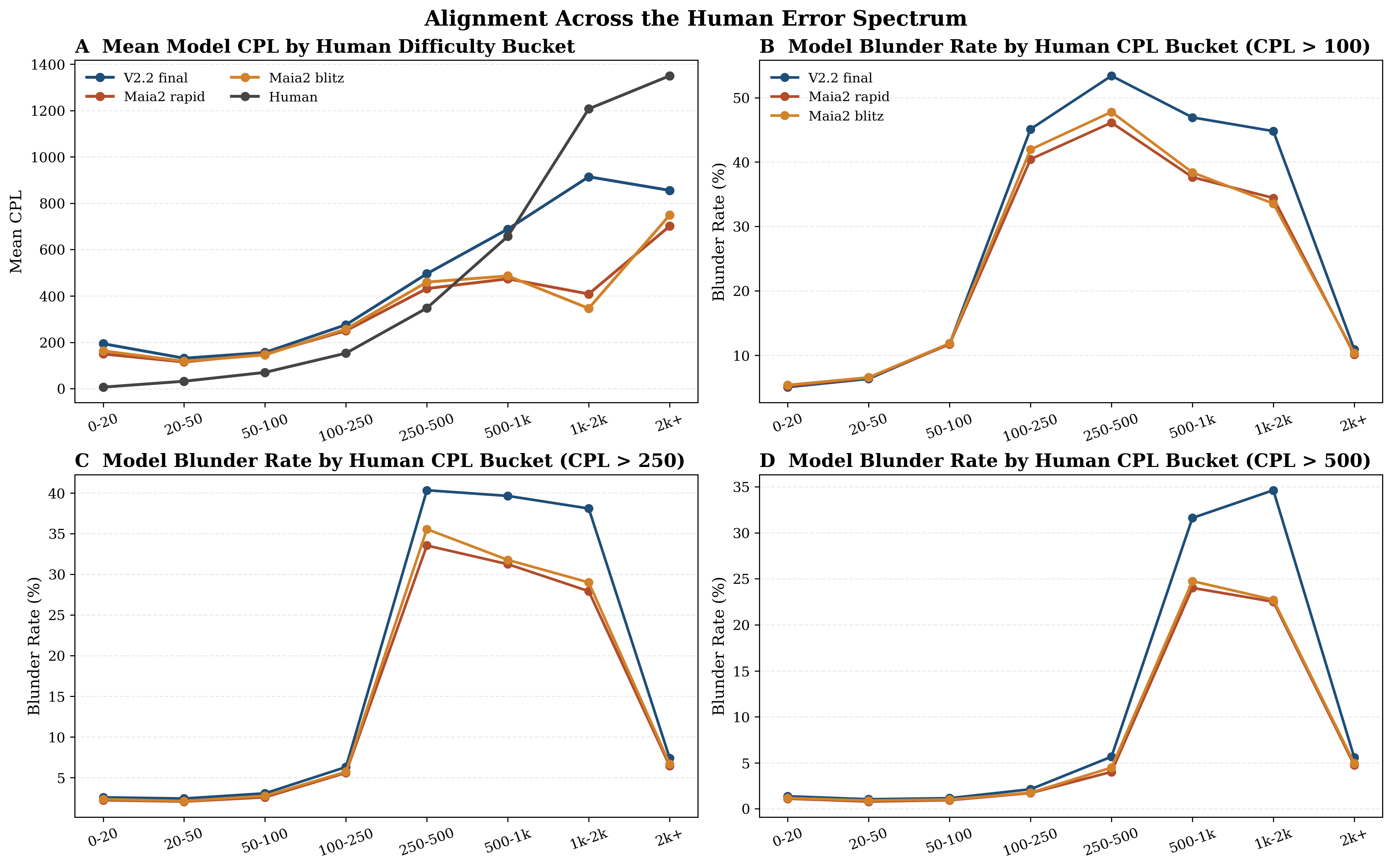}
\caption{Model error across the human difficulty spectrum. Panel~A: mean model CPL by human CPL bucket. V2.2's error profile tracks human difficulty more closely than Maia-2, especially in the 100--500 CPL range where human-like errors concentrate.}
\label{fig:cpl_alignment}
\end{figure}

The CPL alignment analysis (\Cref{fig:cpl_alignment}) further shows that V2.2's mean CPL rises more steeply on positions where humans make larger errors, indicating that its error distribution is shaped by position difficulty in a pattern closer to human cognition.
This is a distinct form of human-likeness from move prediction accuracy: it measures not just \emph{what} the model gets right, but whether it gets \emph{the same things wrong}.

\subsection{History-dependent draw decisions}\label{sec:repetition}

A central structural advantage of sequence models over position-based models is access to full game history.
We test whether our model uses this information with a controlled experiment on \emph{threefold repetition}---a setting where the same position recurs and the player must decide whether to repeat (accepting a draw) or deviate (continuing to play).

\subsubsection{Experimental design}\label{sec:rep_design}

We collect repeated-position decision points from self-play games and group them by Stockfish evaluation into five buckets: big advantage ($>$+200\,cp), advantage (+50 to +200\,cp), equal ($\pm$50\,cp), disadvantage ($-50$ to $-200$\,cp), and big disadvantage ($<$$-200$\,cp).
For each decision point, we measure the probability that the model chooses the repetition-continuing move.

Crucially, we compare two input conditions:
\begin{itemize}[nosep]
  \item \textbf{Full history}: the complete move sequence leading to the position.
  \item \textbf{History stripped}: the repeated trajectory is removed while the current static position is preserved. This isolates the effect of sequential context from the effect of the board state itself.
\end{itemize}

\subsubsection{Results}\label{sec:rep_results}

\begin{figure}[htbp]
\centering
\includegraphics[width=0.9\columnwidth]{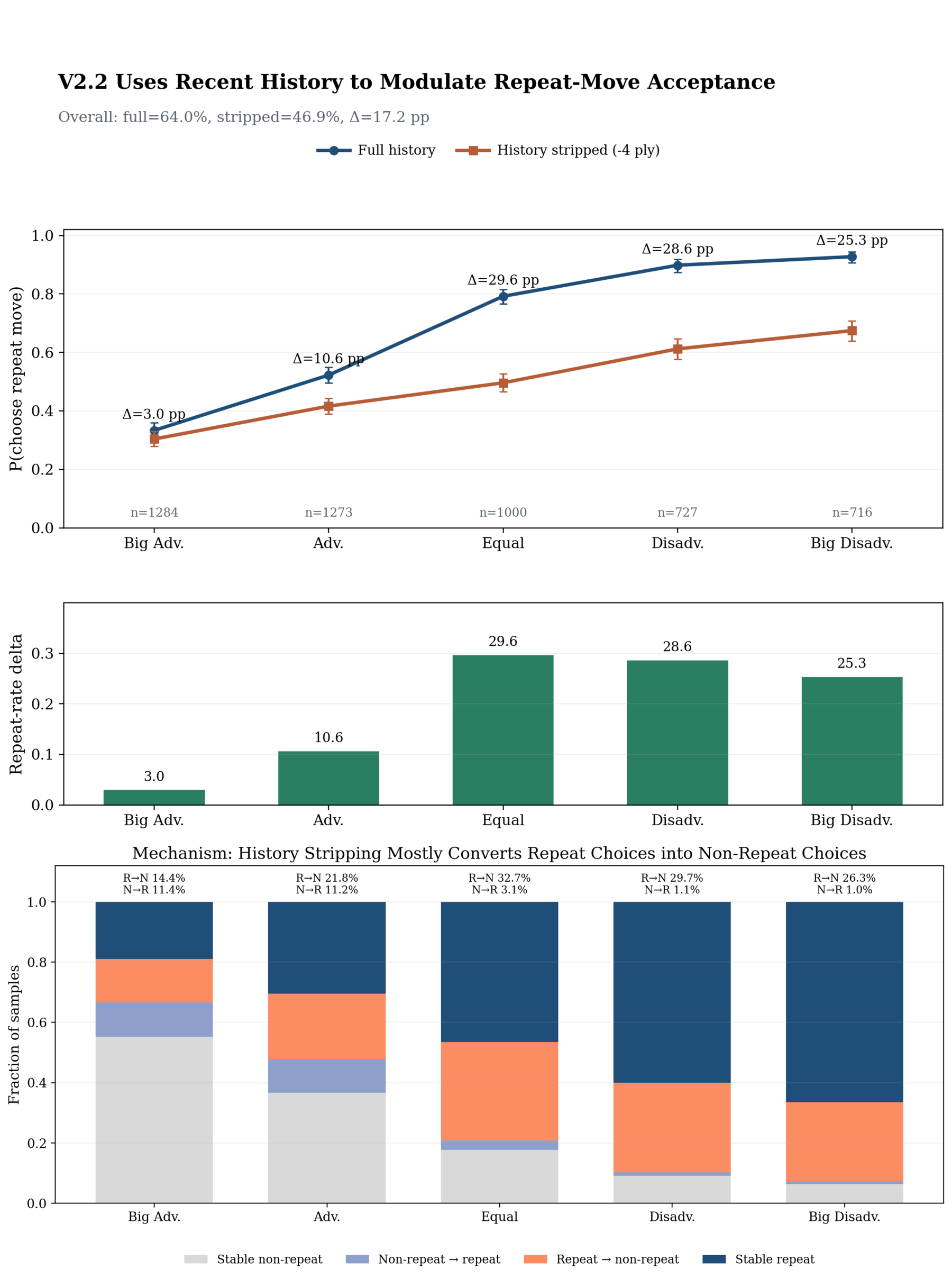}
\caption{Repetition behavior and history-control test. Panel~A: repetition preference across evaluation buckets under full-history and history-stripped inputs. Panel~B: fraction of decisions that flip when history is removed.}
\label{fig:repetition}
\end{figure}

Two findings emerge (\Cref{fig:repetition}):

\paragraph{Context-sensitive risk management.}
The model is less willing to repeat in favorable positions and more willing when equal or worse.
Repetition preference is structured by positional evaluation rather than uniform across positions---qualitatively matching human behavior, where strong players avoid draws when ahead and seek them when under pressure.

\paragraph{History dependence.}
The history-control intervention materially changes repetition behavior: a nontrivial fraction of decisions flip between repeat and non-repeat when history is stripped.
The model is not merely reacting to the current board; it is using the preceding move sequence to decide whether a draw is acceptable.

\subsubsection{Interpretation}\label{sec:rep_interp}

This experiment functions as a \emph{mechanism-level} test rather than an ordinary ablation.
The intervention holds the local board state fixed while selectively removing trajectory information.
The fact that repetition choices change under this intervention demonstrates that the model's draw behavior depends on sequential context---information that FEN-only inputs do not contain.
Position-based models cannot exhibit this behavior by construction, regardless of their accuracy on static move prediction.

\subsection{Deployment check: Lichess play}\label{sec:anticheat}

As a deployment sanity check, we ran V2.2 on Lichess under a standard account.\footnote{Lichess account: \texttt{PineappleDrag}, \url{https://lichess.org/@/PineappleDrag}.}
Over 253 rated bullet games, the model reached a rating of 2570 (top~1\% of active bullet players) with a record of 149 wins, 74 losses, and 30 draws.
The account was not flagged by Lichess's automated anti-cheating system, which monitors statistical anomalies in move timing, accuracy patterns, and engine correlation.

This result is consistent with human-like move and timing patterns, though we note that Lichess's system is primarily tuned to detect engine-assisted play (Stockfish correlation) rather than neural-network play.
It should therefore be read as a deployment consistency check rather than a definitive test of human-likeness.

\FloatBarrier
\section{Discussion}\label{sec:discussion}

\subsection{Supervised fine-tuning on individual style}\label{sec:sft}

Our preliminary SFT experiments produced contradictory results.
On the 28\,M base model (V1.0), five separate SFT experiments on high-quality datasets (Lichess Bullet 2600+, Blitz 2500+, mixtures, TWIC OTB data) all produced models weaker than the untuned base.
On the 120\,M final model (V2.2), fine-tuning on 60,000 chess.com games by Hikaru Nakamura yielded a meaningful improvement of ${\sim}50$--$100$ Elo.

Two hypotheses may explain this discrepancy:
\begin{itemize}[nosep]
  \item \textbf{Capacity hypothesis.} The 28\,M model lacks redundancy; fine-tuning 50\% of layers disrupts tracking--decision coupling. The 120\,M model absorbs adaptation without such disruption.
  \item \textbf{Data specificity hypothesis.} Individual-style data (Hikaru) provides a coherent distributional shift; aggregated high-Elo data provides a noisy mixture.
\end{itemize}
Designing the appropriate cross-controlled experiments is left for future work.

\subsection{Sequence models versus position models}\label{sec:seq_vs_pos}

Two negative results sharpen interpretation.
The short-history variant V1.2 (truncated context) raises the illegal rate from 1.06\% to 2.60\% and degrades accuracy across all phases, showing that a short suffix is insufficient for board reconstruction.
The pure FEN Transformer V4.0 is decisively weaker than the sequence baseline (loses 0--8--2), showing that explicit board input without spatial inductive bias does not automatically help.

These results suggest a division of advantages: position-based models with convolutional structure (\eg, Maia) are parameter-efficient for static move prediction; sequence models are structurally advantaged for trajectory-dependent behavior---style continuity, repetition awareness, and game-level coherence.

\subsection{Applicability to other sequential games}\label{sec:generalization}

Our framework rests on three structural conditions:
(i)~a hidden state exists behind the action sequence,
(ii)~the state space grows exponentially with depth, and
(iii)~data quality and diversity are anti-correlated across skill levels.
These conditions hold across many sequential games---Go, Shogi, Bridge, Poker---with varying parameters.

We offer a testable prediction: in any sequential game where a model is trained on move sequences spanning skill levels, filtering low-skill data will degrade state tracking.
The most accessible validation would use Othello, where existing frameworks \citep{li2023emergent} provide both the training pipeline and linear probe methodology.

\subsection{Limitations}\label{sec:limitations}

\begin{itemize}[nosep]
  \item \textbf{Single domain.} All experiments are on chess; cross-domain validation is pending.
  \item \textbf{Coverage-decay simplifications.} The $t^{*}$ formula assumes uniform branching and a homogeneous support threshold (\Cref{sec:coverage} discusses both limitations explicitly). A phase-dependent branching model and position-complexity-aware thresholds would improve precision but are left for future work.
  \item \textbf{Sweet-spot resolution.} Three weighting intensities ($r = 1, 20, 200$) establish that a sweet spot exists and that both under- and over-weighting are suboptimal, but do not resolve the precise shape of $P(r)$. Since the goal of this work is to demonstrate that weighting is effective and that excess weighting is harmful---not to find the globally optimal $r$---we leave finer-grained sweeps for future work.
  \item \textbf{Two model sizes.} Scaling claims rest on 28\,M and 120\,M; intermediate and larger scales would strengthen them.
  \item \textbf{Evaluation breadth.} Primary evaluation uses Lichess bullet Elo, supplemented by offline metrics. Longer time controls and controlled human experiments would strengthen the claims.
  \item \textbf{Native-input Maia-2 comparison.} The V2.2 vs.\ Maia-2 evaluation compares models under their native input representations (move sequence vs.\ FEN), so it measures the combined effect of model and input information rather than model quality in isolation. A controlled comparison would require training a Maia-2-scale model on move sequences.
\end{itemize}

Despite these limitations, the core contributions---the dual-capability bottleneck, the weighting resolution, and the degeneration analysis---are each supported by multiple independent lines of evidence.

\FloatBarrier
\section{Conclusion}\label{sec:conclusion}

We have shown that training a human-like chess engine from move sequences alone requires balancing two capabilities---state tracking and decision quality---that impose contradictory demands on training data.
The bottleneck $P \le \min(T, Q)$ explains why filtering low-rated data, the intuitive approach, consistently fails: it improves decisions at the cost of tracking, and tracking failure is catastrophic.

Elo-weighted training resolves this paradox by reducing the gradient contribution of low-rated games without removing the positions they generate.
A sweet spot exists: linear weighting ($r{=}20$) balances the two capabilities, while excessive weighting recreates the filtering problem.
A complete $2 \times 2$ factorial ablation supports complementary roles for scaling (tracking) and weighting (decisions), with the combination producing the strongest practical-play gains.
The resulting model is, to our knowledge, the strongest reported pure move-sequence chess system---reaching Lichess bullet 2570 without search, board representations, or hand-crafted chess knowledge.
Under native-input evaluation, it outperforms Maia-2 on human move prediction by 5 percentage points, and a controlled history-removal experiment demonstrates trajectory-dependent draw decisions that FEN-only models structurally cannot produce.

The broader lesson extends beyond chess.
In any autoregressive system where performance depends on multiple capabilities trained from a shared data stream with opposing requirements, na\"ive data curation that optimizes one capability can silently degrade another.
Aggregate training metrics may improve even as practical performance worsens.
The solution lies in \emph{soft} modulation of data influence---preserving signal diversity while shaping gradient emphasis---guided by explicit reasoning about capability bottlenecks.

\bibliographystyle{plainnat}

\end{document}